\documentclass[accepted]{uai2025} % after acceptance, for a revised version; 
\usepackage[american]{babel}
\usepackage{natbib} % has a nice set of citation styles and commands
\usepackage{mathtools} % amsmath with fixes and additions

\usepackage{algorithm} 
\usepackage{algorithmic}
\usepackage{booktabs} % commands to create good-looking tables
\usepackage{tikz} % nice language for creating drawings and diagrams

\usepackage{amsmath,amsfonts,bm}

\def\eqref#1{equation~\ref{#1}}
\def\1{\bm{1}}

\def\vtheta{{\bm{\theta}}}

\def\vl{{\bm{l}}}

\def\vr{{\bm{r}}}

\def\vy{{\bm{y}}}

\def\mA{{\bm{A}}}

\def\mC{{\bm{C}}}

\def\mE{{\bm{E}}}

\def\mR{{\bm{R}}}

\def\mV{{\bm{V}}}
\def\mW{{\bm{W}}}
\def\mX{{\bm{X}}}

\def\mZ{{\bm{Z}}}

\DeclareMathAlphabet{\mathsfit}{\encodingdefault}{\sfdefault}{m}{sl}
\SetMathAlphabet{\mathsfit}{bold}{\encodingdefault}{\sfdefault}{bx}{n}

\def\gG{{\mathcal{G}}}

\def\sV{{\mathbb{V}}}

\newcommand{\R}{\mathbb{R}}

\newcommand{\std}[1]{^{\scriptstyle{\pm#1}}}

\usepackage{hyperref}
\usepackage{url}

\usepackage{microtype}
\usepackage{graphicx}
\usepackage{subfigure}
\usepackage{booktabs}

\def \Wval {W^{\textrm{val}}}

\def \Wcalib {W^{\textrm{calib}}}

\def \loss {\mathcal{L}}

\usepackage{url} 
\usepackage{graphicx} 
\usepackage{amsmath}
\usepackage{amsfonts}
\usepackage{multirow}

\usepackage{adjustbox}
\usepackage{booktabs} 
\usepackage{mathtools}
\usepackage{soul}
\usepackage{bbm}
\usepackage{amsmath}
\usepackage{amssymb}
\usepackage{amsthm}
\usepackage{adjustbox}
\usepackage{booktabs}
\usepackage{ragged2e}
\usepackage{amsmath}
\usepackage{svg}
\usepackage{adjustbox}
\usepackage{booktabs}

\usepackage{amsmath}
\usepackage{amssymb}
\usepackage{mathtools}
\usepackage{amsthm}

\title{Residual Reweighted Conformal Prediction for Graph Neural Networks}

\usepackage[american]{babel}

\usepackage{graphicx}
\usepackage{amsmath}
\usepackage{hyperref}

\newcommand{\equalcontrib}{\textsuperscript{*}}
\newcommand{\corrauthor}{\textsuperscript{\dag}}

\author[1]{Zheng Zhang\equalcontrib}
\author[2,3]{Jie Bao\equalcontrib}
\author[4]{Zhixin Zhou}
\author[5]{Nicolo Colombo}
\author[6]{Lixin Cheng}
\author[1,2]{\href{mailto:<ruiluo@cityu.edu.hk>?Subject=Residual Reweighted Conformal Prediction for Graph Neural Network}{Rui Luo\corrauthor}{}}

\affil[1]{Department of System Engineering, City University of Hong Kong, China}
\affil[2]{Chengdu Research Institute, City University of Hong Kong, China}
\affil[3]{Faculty of Electronic Information Engineering, Huaiyin Institute of Technology, China}
\affil[4]{Alpha Benito Research, USA}
\affil[5]{Centre for Intelligent Systems, Royal Holloway, University of London, UK}
\affil[6]{Shenzhen People's Hospital, China}

\begin{document}
\maketitle
\begingroup
\renewcommand{\thefootnote}{\fnsymbol{footnote}}
\footnotetext[1]{These authors contributed equally to this work.}
\footnotetext[2]{Corresponding author: \texttt{ruiluo@cityu.edu.hk}}
\endgroup

\begin{abstract}
Graph Neural Networks (GNNs) excel at modeling relational data but face significant challenges in high-stakes domains due to unquantified uncertainty. Conformal prediction (CP) offers statistical coverage guarantees, but existing methods often produce overly conservative prediction intervals that fail to account for graph heteroscedasticity and structural biases. While residual reweighting CP variants address some of these limitations, they neglect graph topology, cluster-specific uncertainties, and risk data leakage by reusing training sets. To address these issues, we propose Residual Reweighted GNN (RR-GNN), a framework designed to generate minimal prediction sets with provable marginal coverage guarantees.

RR-GNN introduces three major innovations to enhance prediction performance. First, it employs Graph-Structured Mondrian CP to partition nodes or edges into communities based on topological features, ensuring cluster-conditional coverage that reflects heterogeneity. Second, it uses Residual-Adaptive Nonconformity Scores by training a secondary GNN on a held-out calibration set to estimate task-specific residuals, dynamically adjusting prediction intervals according to node or edge uncertainty. Third, it adopts a Cross-Training Protocol, which alternates the optimization of the primary GNN and the residual predictor to prevent information leakage while maintaining graph dependencies. We validate RR-GNN on 15 real-world graphs across diverse tasks, including node classification, regression, and edge weight prediction. Compared to CP baselines, RR-GNN achieves improved efficiency over state-of-the-art methods, with no loss of coverage.
\end{abstract}

\section{Introduction}
Graph Neural Networks (GNNs) have achieved state-of-the-art performance on graph-structured data across various applications like recommendation systems, knowledge graphs, and molecular modeling \cite{lam2023learning, li2022graph, wu2022graph}. As GNNs are increasingly applied in high-stakes areas such as healthcare and autonomous systems, accurately assessing prediction uncertainty becomes paramount. A common approach to predicting uncertainty is to construct prediction intervals that capture the probability of true outcomes. While several methods of predicting uncertainty have been explored \cite{hsu2022makes, zhang2020mix, lakshminarayanan2017simple}, they typically lack rigorous theoretical guarantees on interval validity \cite{wang2021confident}. Improving uncertainty quantification for GNNs with probabilistic guarantees is critical to ensuring their safe, trusted application in real-world settings. 

Previous residual reweight methods~\cite{papadopoulos2008normalized} often relied on fixed weight loss functions and importance weighting
techniques, which do not adequately capture heteroscedasticity, leading to inaccurate quantification of prediction
errors across diverse samples. Guan~\cite{guan2023localized} proposed Localized Conformal Prediction (LCP), and Han et al. proposed
Split Localized Conformal Prediction (SLCP)~\cite{han2022split} which uses kernel-based weights. However, kernel methods
are well-known to suffer from the curse of dimensionality, making them less effective in handling complex data,
particularly structured data such as graphs. Previous graph-based methods often relied on simplistic partitioning
techniques that ignored the complex topological structure between nodes and edges~\cite{clarkson2023distribution}, thereby lacking the
capability to perform the samplewise normalization.
Conformal Prediction (CP) is a machine learning framework for uncertainty quantification that constructs prediction intervals for any underlying point predictor in a theoretically valid manner \cite{vovk2005algorithmic}. Due to its principled formulation, rigorous guarantees, and distribution-free nature, CP has enabled uncertainty estimation across diverse applications, including computer vision \cite{angelopoulos2020uncertainty, bates2021distribution}, causal inference \cite{lei2021conformal, jin2023sensitivity, yin2024conformal}, time series \cite{gibbs2021adaptive, zaffran2022adaptive}, and drug discovery \cite{jin2023selection}. CP leverages a "calibration" dataset to output prediction sets for new test samples that provably cover the true outcome with at least 1 - $\alpha$ probability, where $\alpha$ is a user-specified error tolerance. CP is based on a nonconformity measure/score that measures the dissimilarity between a data point and others, reflecting disagreements according to the algorithm's feature-relationship assumptions. Crucially, each nonconformity score can represent a single algorithm, by defining a distinct CP predictor \cite{papadopoulos2008normalized}. Additionally, adding a Residual-Reweighting (RR) factor can refine prediction intervals \cite{papadopoulos2011regression, lei2018distribution}. By assigning weights to errors covariately, RR helps mitigate heteroscedasticity impacts on accuracy and reliability. Overall, carefully constructing the nonconformity measure and incorporating RR are pivotal to prediction performance. 

Previous residual reweight methods often relied on fixed-weight loss functions and importance weighting techniques to address sample imbalance in various applications. However, these approaches have significant limitations: they often overlook relationships between samples, fail to leverage the topological structure of the data, and inadequately analyze prediction residuals when dynamically adjusting sample weights. Existing methods, such as normalized nonconformity functions \cite{johansson2014regression,kath2021conformal} and local reweighted conformal methods \cite{papadopoulos2008normalized}, involve training separate models to predict errors but struggle to capture the heteroscedasticity of the data. Similarly, prior graph-based methods often relied on simplistic partitioning techniques that ignored complex relationships between nodes and edges \cite{aggarwaldata}. Before the Cross-Training Protocol, machine learning models, including GNNs, faced challenges with information leakage during training \cite{vepakomma2020nopeek}. Simultaneous optimization of multiple models on the same dataset often caused one model’s learning to adversely affect the other, leading to biased predictions and reduced performance \cite{kapoor2023leakage,hitaj2017deep}.

We propose a novel residual-reweighted nonconformity measure for graph neural networks (RR-GNN), which optimizes model performance by independently predicting expected accuracy. RR-GNN consists of two GNNs: the Conformal GNN, which generates predictions based on input features and true labels, and the Residual GNN, trained on prediction residuals to calibrate outputs.

We present the key contributions of this work:
\begin{itemize}
    \item We propose a novel framework that integrates conformal prediction with GNNs to enhance uncertainty quantification in graph-structured data.
    \item To address the heteroscedasticity of graph data, we design a novel nonconformity score that reweights residuals using a separately trained GNN.
    \item We introduce a graph-based Mondrian CP, where nodes are clustered based on graph structure, enabling fine-grained, context-aware prediction intervals \cite{aggarwaldata}.
    \item To prevent information leakage between the primary GNN and residual-prediction GNN, we develop a cross-training strategy where models iteratively update each other. This ensures independence between calibration and training data—a critical requirement for CP validity—while maintaining model performance \cite{vepakomma2020nopeek,kapoor2023leakage,hitaj2017deep}.
\end{itemize}

\section{Related Work}

Conformal prediction (CP)~\cite{vovk2005algorithmic} is a methodology designed to generate prediction regions for variables of interest, facilitating the estimation of model uncertainty by providing prediction sets rather than point estimates. CP has been successfully applied to both classification~\cite{luo2024trustworthy,luo2024entropy,luo2024weighted} and regression tasks~\cite{luo2025threshold,luo2025volume}. Its flexibility allows adaptation to various real-world scenarios, including segmentation~\cite{luo2025conditional}, games~\cite{luo2024game, bao2025enhancing}, time-series forecasting~\cite{su2024adaptive}, and graph-based applications~\cite{luo2023anomalous, tang2025enhanced, luo2025conformalized, wang2025enhancing, luo2025conformal}.

Graph Neural Networks (GNNs) have become foundational models for learning from graph-structured data. Kipf et al.~\cite{kipf2016semi} introduced a seminal unsupervised GNN for learning low-dimensional embeddings. Cai et al.~\cite{cai2021line} proposed the Line Graph Neural Network (LGNN), which reformulates link prediction as a node classification problem in the line graph. Kollias et al.~\cite{kollias2022directed} developed DiGAE, a directed graph auto-encoder with parameterized message passing for node classification and link prediction. GNNs are also adaptable to regression tasks \cite{luo2023you} by modifying the output layer and loss functions, which has potential applications in network segregation prediction and control \cite{luo2021echo, krishnamurthy2021segregation, luo2022controlling, luo2022mitigating, luo2023frechet}.

For heterogeneous graphs, specialized architectures have been proposed. Wang et al.~\cite{wang2019heterogeneous} introduced the Heterogeneous Graph Attention Network (HAN), which employs hierarchical attention to capture both meta-path-based and semantic-level importance, achieving state-of-the-art results. Iyer et al.~\cite{iyer2021bi} presented the Bi-Level Attention GNN (BAGNN), utilizing a bi-level attention mechanism to learn complex relationships in heterogeneous data.

While existing GNN approaches focus primarily on achieving high prediction accuracy, they often lack flexibility, adaptability, and mechanisms for uncertainty quantification. Conformal prediction addresses these gaps by offering prediction intervals and enhanced generalization. Notably, Huang et al.~\cite{huang2024uncertainty} extended CP to GNNs with CF-GNN, improving uncertainty quantification. Robustness in these frameworks has been further enhanced by importance weighting schemes. Guo et al.~\cite{guo2017calibration} introduced a causal inference-driven weighting technique, whereas Volpi et al.~\cite{volpi2018generalizing} proposed a distribution matching-based strategy to mitigate distribution shifts. 

Despite these advancements, rigorous guarantees on coverage and reliability for uncertainty quantification in GNNs remain an open challenge~\cite{bhagat2011node}.

\begin{figure}[ht]
\centerline{\includegraphics[height=9.2cm,width=0.45\textwidth,clip=]{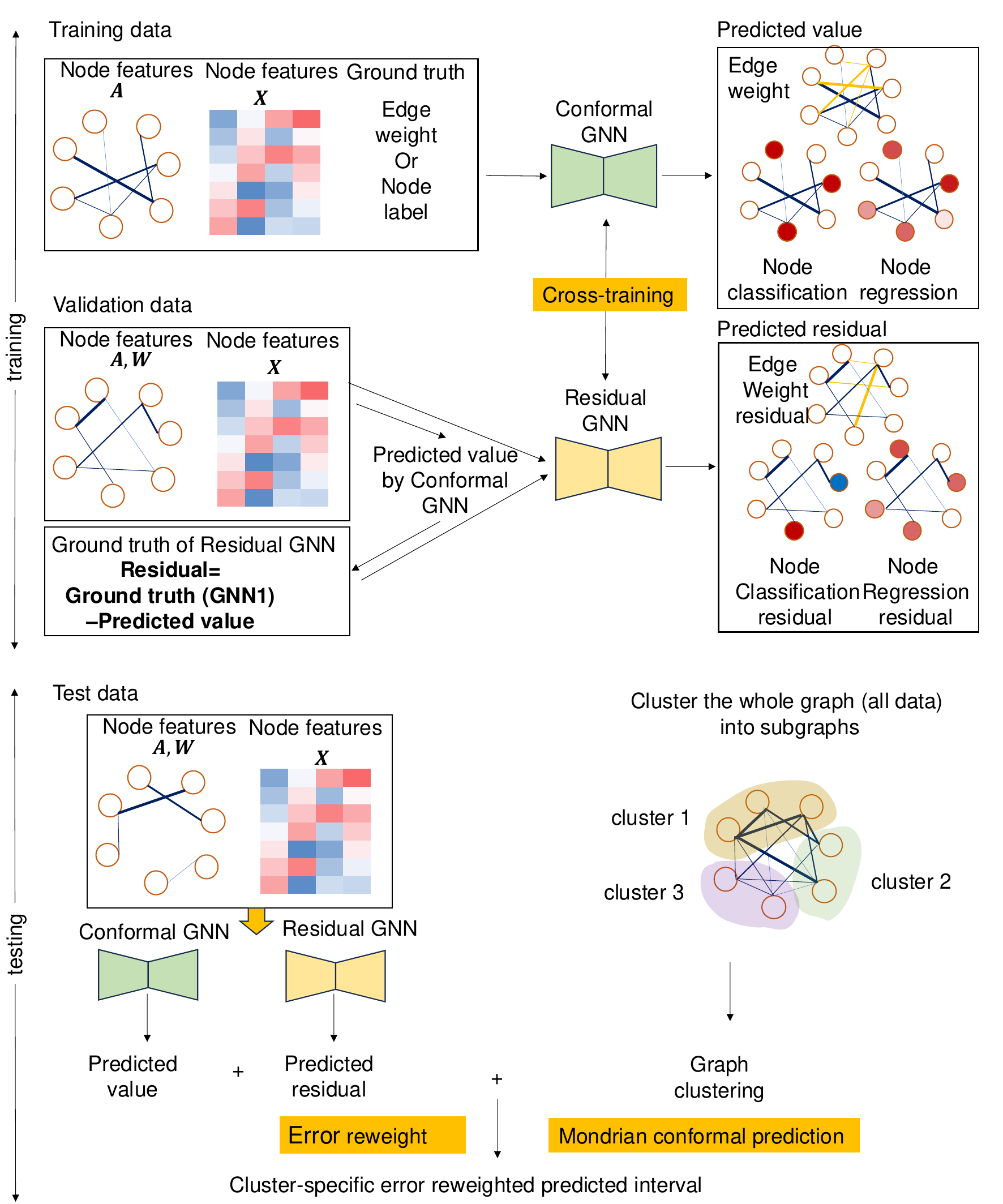}}

\caption{The RR-GNN pipeline consists of three parts: 1) Conformal GNN Training: A GNN is trained for edge weight prediction, node regression, or classification, outputting prediction intervals with uncertainty quantification and marginal coverage guarantees; 2) Residual GNN Training: A separate Residual GNN is trained on validation data to predict residuals between true values and Conformal GNN predictions; 3) Residual Reweighting: Prediction intervals from the Conformal GNN and residual weights from the Residual GNN are integrated, and Mondrian conformal prediction incorporates graph-structured clustering. The Conformal and Residual GNNs are trained alternately, with the Residual GNN correcting the Conformal GNN’s final predictions.}
\label{fig: modelpipeline}
\end{figure}

\section{Methodology}\label{sec: m}
RR-GNN combines conformal prediction with a GNN-based framework to effectively address graph-structured data tasks. The model accepts a graph as input, either undirected or directed and yields predictions for edge weights or node features as output. RR-GNN
provides prediction intervals instead of a single point
value, which can better capture the uncertainty in the
data. For new test samples, the probability of the predicted interval would cover the true outcome with at least 1 - $\alpha$ with rigorous theoretical guarantees.

 RR-GNN performs graph-based Mondrian CP, in which the input graph is clustered to subnetworks to capture the community structure of the graph. It begins by training a predictive model on a training dataset named Conformal GNN, which gives the main prediction according to the tasks. Next, RR-GNN trains a Residual GNN model based on the validation set to predict the residual of Conformal GNN, which is next used to calculate prediction errors used as a reweight factor to establish nonconformity scores, measuring
how unusual the data point looks relative to previous examples. After determining a significance level based on the desired confidence, RR-GNN generates a cluster-specific fixed prediction interval based on the distribution of nonconformity scores in the calibration set of each cluster. The predicted interval of test data is given by the combination of predicted point estimation, residual prediction and the cluster-specific fixed interval. 
Notably, throughout this process, each dataset is divided into four distinct parts: training data, validation data, calibration data, and test data, with only the test data lacking labels. Algorithm~\ref{closs_train} and Figure~\ref{fig: modelpipeline} show the implementation details of training the main and residual models crossly simultaneously~\cite{cowell2006alternative,peste2021ac}. The Conformal GNN model is
trained using a cross-training process with the Residual GNN model. %The efficiency of cross-training of two residual model can be shown in \ref{prop: crosstraining2}.

We clarify the notation here. The input is a graph,  $\displaystyle \gG=(\displaystyle \sV,\displaystyle \mE)$, with node set $\displaystyle \sV$ and edge set $\displaystyle \mE \subseteq \displaystyle \sV \times \displaystyle \sV$.
Assume the graph has $n$ nodes with $m$ features.
Let $\displaystyle \mX \in \displaystyle \R^{n\times m}$ be the node feature matrix, and $\mX_{i,:} \in \displaystyle \R^{m}$ be the feature vector of the $i$th node. 
The binary adjacency matrix of $\displaystyle \gG$, $ \displaystyle \mA \in \{0, 1\}^{n\times n}$, encodes the binary (unweighted) connecting structure of the graph. For the weighted graph, we use $ \displaystyle \mW \in \displaystyle \R^{n\times n}$ to represent the weighted adjacency matrix.

In RR-GNN, taking the weighted graph as an example. We partition the edge set $\mE$ into three disjoint subsets: $\mE^{train}$, $\mE^{val}$, $\mE^{calib}$ and $\mE^{test}$, while satisfying $\mE = \mE^{train} \cup \mE^{val} \cup \mE^{calib} \cup \mE^{test}$. 
We define
\begin{equation}\label{eq: weighted adj train}
\textbf{W}^{train} =
\begin{cases}
W_{ij}, & \textrm{if } (i, j) \in \mE^{train}; \\
\delta, & \textrm{if } (i, j) \in \mE^{val} \cup \mE^{calib} \cup \mE^{test}; \\
0, & \textrm{otherwise},
\end{cases}
\end{equation}
where $\delta > 0$ is a small positive constant to represent a minimal edge weight.
The Conformal GNN is represented by:
\begin{equation}
\label{eq: CGNN}
    [\hat{\vy},\hat{\vy}^{\alpha/2},\hat{\vy}^{1-\alpha/2}] = g_{\vtheta_{1}}\left( \displaystyle \mW^{train}, \displaystyle \mX \right) ,
\end{equation}
where $y$ is the label for a specific task; $g_{\vtheta_{1}}$ represents the GNN-based model mapping the input to the label; $\hat{\vy}^{\alpha/2}$ and $\hat{\vy}^{1-\alpha/2}$ is the predicted $\alpha/2$ and $1-\alpha/2$ quantile of the label. The Residual GNN is represented by the following equation:
\begin{equation}
\label{eq: RGAE0}
    \hat{\mR} = g_{\vtheta_{2}}\left( \displaystyle \mW^{val}, \displaystyle \mX \right) ,
\end{equation}
where $\hat{\mR}$ is the predicted residual of Conformal GNN prediction; $g_{\vtheta_{2}}$ represents the GNN-based model mapping the input to the residual. 
Then, we will predict residuals for the calibration set and calculate the nonconformity score
\begin{equation}
\label{eq: ncs}
    \mV = s\left(g_{\vtheta_{1}}\left( \displaystyle \mW^{calib}, \displaystyle \mX \right),g_{\vtheta_{2}}\left( \displaystyle \mW^{calib}, \displaystyle \mX \right) \right),
\end{equation}
where $s(.)$ is the nonconformity function. We will perform graph-based partition: the nodes in the graph $G=(V, E)$ are clustered into $K$ groups $\mathcal{C}_1, \mathcal{C}_2, \ldots, \mathcal{C}_K$ using Louvain clustering, where $G_{(m)}=(V_{(m)}, E_{(m)})$ is the subgraph for the cluster $m$.
Next, we will get an interval factor for each cluster, $d_{(m)}$, which is a quantile of nonconformity scores according to the significant level.
The predicted interval for a test data point in cluster m is 
\begin{multline}
\label{eq: itv}
    \mC_{(m)} = [l\left(g_{\vtheta_{1}}\left( \displaystyle \mW^{test}, \displaystyle \mX \right),g_{\vtheta_{2}}\left( \displaystyle \mW^{test}, \displaystyle \mX \right) , d_{(m)}\right),\\
    u\left(g_{\vtheta_{1}}\left( \displaystyle \mW^{test}, \displaystyle \mX \right),g_{\vtheta_{2}}\left( \displaystyle \mW^{test}, \displaystyle \mX \right) , d_{(m)} \right) ],
\end{multline}
where $l(.)$ and $u(.)$ represent the lower and upper bound of the prediction interval.

\begin{algorithm}[tb]
   \caption{Crossly-Training Algorithm for RR-GNN}
   \label{closs_train}
\begin{algorithmic}
   \STATE {\bfseries Input:} Model $C_0$ (Conformal GNN) weights $\vtheta_1 \in \displaystyle \R^N$, Residual Model $C_1$ (Residual GNN) weights $\vtheta_2 \in \displaystyle \R^N$, loop limit $n$
   \FOR{$i=1$ {\bfseries to} $n$ {\bfseries with step} $1$}
      \IF{$i$ is odd}
         \STATE Train Model $C_0$ with gradients and update $\vtheta_1$ using the training data.
      \ELSE
         \STATE Get the residual from Model $C_0$ as the label based on validation data.
         \STATE Train Residual Model $C_1$ with gradients and update $\vtheta_2$ based on validation data.
      \ENDIF
   \ENDFOR
\end{algorithmic}
\end{algorithm}

\subsection{Node Regression/Classification}
For the node regression/classification tasks, we partition the node set $\mV$ into three disjoint subsets: $\mV^{train}$, $\mV^{val}$, $\mV^{calib}$ and $\mV^{test}$, while satisfying $\mV = \mV^{train} \cup \mV^{val} \cup \mV^{calib} \cup \mV^{test}$. 
We define the label of node $\vy$ as:
\begin{equation}\label{eq: weighted adj train2}
\vy^{train} =
\begin{cases}
y_{i}, & \textrm{if } i \in \mV^{train}; \\
0, & \textrm{otherwise},
\end{cases}
\end{equation}
We use the training label $\vy^{train}$ and validation label $\vy^{val}$  to train the Conformal GNN and get the parameters $\vtheta_{1}$ and $\vtheta_{2}$, respectively.
Then, we will predict residuals for the calibration set and calculate the nonconformity score for each node from the calibration set.
\begin{equation}
\label{eq: ncs2}
    \mV^{calib} = s\left(g_{\vtheta_{1}}\left( \displaystyle \mW, \displaystyle \mX \right),g_{\vtheta_{2}}\left( \displaystyle \mW, \displaystyle \mX \right),\sV^{calib} \right),
\end{equation}
where the nonconformity score is calculated only on the calibration set.
Next, we will get an interval factor, $d$, which is a quantile of nonconformity scores according to the significant level. The predicted interval for a test data point is 
\begin{multline}
\label{eq: itv22}
    \mC =
    [l(g_{\vtheta_{1}}\left( \displaystyle \mW, \displaystyle \mX \right),g_{\vtheta_{2}}\left( \displaystyle \mW, \displaystyle \mX \right) , d,\sV^{test}), \\
    u(g_{\vtheta_{1}}\left( \displaystyle \mW, \displaystyle \mX \right),g_{\vtheta_{2}}\left( \displaystyle \mW, \displaystyle \mX \right) , d,\sV^{test}) ],
\end{multline}
where $l(.)$ and $u(.)$ represent the lower and upper bound of the prediction interval, which are calculated based on the predicted node label of the test set.

\subsection{RR-GNN on Edge Weight Prediction}
\subsubsection{Conformal GAE for Edge Weight Prediction } \label{subsec: GAE}
 The GAE \cite{kipf2016variational,ahn2021variational} is used for edge weight prediction tasks by learning node embeddings across various types of graphs, including directed graphs \cite{kollias2022directed}, weighted graphs \cite{zulaika2022lwp}, and graphs with different edge types \cite{samanta2020nevae}. The edge weight is then given by the similarity of node embeddings. There are two kinds of problem settings in link prediction shown in Figure 1 in appendices material including transductive setting and inductive setting. We focus on the first one. We can see the details in the first part in Figure~\ref{fig: transductive} appendices material. We integrate CP in GAE's framework by making the encoder produce a triple output.
 We use $\displaystyle \mZ, \mZ^{\alpha/2}$, and $\mZ^{1-\alpha/2}\in \displaystyle \R^{n\times d}$ to represent the mean, $\alpha$/2 quantile, and $(1-\alpha / 2)$ quantile of node embedding matrix obtained from a Conformal GAE model. This differs from having three single-output GAE encoders because most network parameters are shared across the three embeddings.
The resulting embedding is 
\begin{equation}
[\mZ,\mZ^{\alpha/2},\mZ^{1-\alpha/2}] = f_{\vtheta}(\displaystyle \mX, \displaystyle \mA),
\label{eq:gae encoder}
\end{equation}
where $f_{\vtheta}$ is the structure of the encoder, and $\vtheta$ is a learnable parameter. Note that the traditional GNN model is applicable because it could generate $d$-dimentional output for each node, which represents node embeddings. Directed GAE designed for the directed graph \cite{kollias2022directed} is more flexible, using separate source and target embeddings, $\mZ=[\displaystyle \mZ^S,\mZ^T]$. As for undirected GAE, $\mZ^S=\mZ^T$. It is similar for $\mZ^{\alpha/2}$ and $\mZ^{1-\alpha/2}$.

We take the directed graph as an example for the following description. We next reconstruct the weighted adjacency matrix from the inner product between node embeddings, which is the Conformal GNN-based model. 
\begin{multline}
     g_{\vtheta_1}(\mX,\mA)=[\hat{\displaystyle \mW}, \hat{ \mW}^{\alpha/2}, \hat{\mW}^{1-\alpha/2}]=  [\mZ^S {( \mZ^T)}^\top,\\
     \mZ^{S,\alpha/2} {( \mZ^{T,\alpha/2})}^\top,\mZ^{S,1-\alpha/2} {( \mZ^{T,1-\alpha/2})}^\top].
\end{multline}
where $\hat{\mW}, \hat{\mW}^{\alpha / 2}$, and $\hat{\mW}^{1-\alpha / 2}$ be the mean, $\alpha / 2$, and (1- $\alpha$/2) quantiles of the edge weights.

The loss function $\mathcal{L}_{\mathrm{Conformal}-\mathrm{GNN}}$ is given by:
\begin{multline}
\mathcal{L}_{\mathrm{GAE}} +
   \sum_{(i, j) \in E^{\text{train}}} \rho_{\alpha / 2}\left(W_{i j}^{\text{train}}, \hat{W}_{i j}^{\alpha / 2}\right) \\
   \quad + \rho_{1-\alpha / 2}\left(W_{i j}^{\text{train}}, \hat{W}_{i j}^{1-\alpha / 2}\right).
\end{multline}

where $\mathcal{L}_{\mathrm{GAE}}$ is the squared error loss defined in (9) The second term is the pinball loss referenced to~\cite{romano2019conformalized,steinwart2011estimating}, defined as
$$
\rho_\alpha(y, \hat{y}):= \begin{cases}\alpha(y-\hat{y}) & \text { if } y>\hat{y} \\ (1-\alpha)(y-\hat{y}) & \text { otherwise }\end{cases}
$$

The first term is added to train the mean estimator, $\hat{\mW}$. 
%Algorithm $\underline{2}$ describes how to obtain the prediction intervals in this setup. Contrary to the CP conformity score (13), the CQR conformity score (18) considers both undercoverage and overcoverage scenarios. 
\begin{equation} \label{eq:Train_GAE}
    \loss_{\textrm{GAE}} = \| \displaystyle \mA^{train} \odot \hat{\displaystyle \mW} -\displaystyle \mW^{train} \|_.    
\end{equation}
%We train the model until convergence and then select the tuning parameters that minimize $\loss_{\textrm{GAE}}$ on the validation set, $\mW^{val}$.

\subsubsection{RR-GNN on Edge Weight Prediction}\label{sec: conformal}

We train a separate Residual GAE model  to predict the error of the edge weight prediction of the $g_{\vtheta_1}$, 
\begin{equation}
\label{eq: RGAE}
    \hat{R}_{ij}^{val} = g_{\vtheta_{2}}\left((i,j); \displaystyle \mA, \displaystyle \mX \right) ,
\end{equation}
where the label is $R^{val}_{ij} = \Wval_{ij} - \hat{W}_{ij}^{val}$, $\hat{W}_{ij}^{val}=g_{\vtheta_{1}}\left((i,j); \displaystyle \mA^{val},\displaystyle \mX \right)$ is the output of conformal GAE, and the RR-GAE is trained by minimizing
\begin{equation} \label{eq: train RGAE}
    \loss_{\textrm{Residual GNN}} = \| \displaystyle \mA^{val} \odot \hat{\displaystyle \mR}^{val} - \displaystyle \mR^{val} \|_F.  
\end{equation}
\hfill

We use the standard deviation of these predictions as a proxy of the residual. More concretely, we propose a new nonconformity score function, which is the interval of predicted edge weight reweighted according to the absolute value of the residual as predicted by the RR-GNN model (\ref{eq: RGAE}).
\begin{equation}\label{eq: Vij}
V^{\textrm{RR}}_{ij} = \max\left \{ \frac{\hat{W}^{\alpha/2}_{ij} - \Wcalib_{ij}}{\big|\hat{R}_{ij}\big|}, \frac{\Wcalib_{ij} - \hat{W}^{1-\alpha/2}_{ij}}{\big|\hat{R}_{ij}\big|}  \right\}, \; 
\end{equation}
\begin{equation}
(i, j) \in \displaystyle \mE^{calib},  
\end{equation}
where $\hat{\mW}^{\alpha/2}$ and $\hat{\mW}^{1-\alpha/2}$ is the predicted edge weight quantile of the Conformal GAE based on the calibration set.  
Let $d^{\textrm{RR}}_{(m)} $ be the $k$-th smallest value in $\{V^{\textrm{RR}}_{ij}|(i,j)\in \mE^{calib}_{(m)}\}$ for the cluster m, where $k=\lceil(n/2 +1)(1-\alpha)\rceil$ where n is size of $\displaystyle \mE^{calib}_{(m)}$. 
The RR prediction intervals for cluster m are:
\begin{equation}\label{eq: interval RR R-GAE}
   C_{ab}^{(m)} =  \Big[ \hat{W}^{\alpha/2}_{ab} - d^{\textrm{RR}}_{(m)} \big|\hat{R}_{ab}\big|,  \;
    \hat{W}^{1-\alpha/2}_{ab} + d^{\textrm{RR}}_{(m)}
    \big|\hat{R}_{ab}\big| \Big], \;
\end{equation}
\begin{equation}    
    (a, b) \in \displaystyle \mE^{test}_{(m)},
\end{equation}
The theoretical guarantees on interval validity can be referenced to~\cite{luo2025conformal}.

\begin{algorithm}[ht]
   \caption{Residual Reweighted Conformalized Graph Neural Network for Edge Weight Prediction}
   \label{alg: CQR}
\begin{algorithmic}[1]
   \STATE {\bfseries Input:} The binary adjacency matrix $\displaystyle \mA \in \{0, 1\}^{n\times n}$, edge weight matrix $\displaystyle \mW \in \displaystyle \R^{n\times n}$, node features $\displaystyle \mX\in \displaystyle \R^{n\times m}$, training edges and weights $\displaystyle \mE^{train}$ and $\displaystyle \mW^{train}$, validation edges and weights $\displaystyle \mE^{val}$ and $\displaystyle \mW^{val}$ (used for training Residual GNN), calibration edges and weights $\displaystyle \mE^{calib}$ and $\displaystyle \mW^{calib}$, and test edges $\displaystyle \mE^{test}$, user-specified error rate $\alpha \in (0,1)$, two GNN models $g_{\vtheta_1}$ and $g_{\vtheta_2}$ with trainable parameters $\vtheta_1$ and $\vtheta_2$.
   \STATE Cluster the whole graph $\mE = \mE^{train} \cup \mE^{val} \cup \mE^{calib} \cup \mE^{test}$ into $K$ clusters using Louvain clustering.
   \STATE Train the models $g_{\vtheta_1}$ and $g_{\vtheta_2}$ with $\mA$, $\mX$, $\displaystyle \mW^{train}$ and $\displaystyle \mW^{val}$ according to Algorithm \ref{closs_train}.
   \STATE Predict the interval $[\hat{W}^{\alpha/2}_{ij},\hat{W}^{1-\alpha/2}_{ij}]$ as the output of $g_{\vtheta_1}$ and the residual $\hat{R}_{ij}$ as the output of $g_{\vtheta_2}$ using the calibration data as input.
   \STATE Compute the nonconformity score $V^{\textrm{RR}}_{ij}$ for the calibration data according to (\ref{eq: Vij}).
   \STATE Compute $d_{(m)}=$ the $k$-th smallest value in $\{V^{\textrm{RR}}_{ij}|(i,j)\in \mE^{calib}_{(m)}\}$, where $k=\lceil(|\displaystyle \mE^{calib}_{(m)}| +1)(1-\alpha)\rceil$.
   \STATE Construct a prediction interval for test edges according to (\ref{eq: interval RR R-GAE}).
   \STATE {\bfseries Output:} Prediction of confidence intervals for the test edges $(a, b) \in \displaystyle \mE^{test}$ with the coverage guarantee according to (\ref{eq: interval RR R-GAE}).
\end{algorithmic}
\end{algorithm}

\begin{table*}[t]
\caption{Performance comparison of the proposed models}
\label{tab:eff_all_models1gsmg}
\centering
\begin{adjustbox}{width=0.85\textwidth}
\begin{tabular}{|l|c|c|c|c|c|c|c|c|}
\toprule

GNN Model On Chicago Data & \multicolumn{2}{c|}{GraphConv} & \multicolumn{2}{c|}{SAGEConv}  & \multicolumn{2}{c|}{GCNConv} & \multicolumn{2}{c|}{GATConv} \\ \cmidrule{1-9}
Score Method-CP  & cover$^x$  & ineff & cover$^x$ & ineff & cover$^x$ & ineff& cover$^x$ & ineff\\\midrule
GAE&$0.7984\std{0.1181}$ 
&$3.6659\std{0.3313}$
&$0.8297\std{0.1264}$
&${3.6350}\std{0.2231}$
&$0.8234\std{0.1213}$ 
&$3.6918\std{0.2454}$
&$0.9524\std{0.0333}$
&$3.3493\std{0.5910}$\\
DiGAE&$0.8081\std{0.1257} $ 
&${3.5721}\std{0.1951}$
&$0.8196\std{0.1215}$
&$3.5978\std{0.1884}$
&$0.8135\std{0.1361} $ 
&$3.5846\std{0.2050}$
&$0.8135\std{0.1319}$
&$3.6346\std{0.2432}$\\
LGNN&$0.9174\std{0.0238}$ 
&$6.7157\std{0.1325}$
&$0.9152\std{0.0256}$
&$6.5865\std{0.1577}$
&$0.9151\std{0.0246}$ 
&$6.5265\std{0.1426}$
&$0.9075\std{0.0618}$
&${6.0679}\std{0.1862}$\\ 
\midrule
Average & 0.8477 & 4.6512 & 0.8548 & 4.5998 & 0.8507 & 4.6010 & 0.8912 & 4.3506 \\
\midrule 
Score Method-CQR  & cover$^x$  & ineff & cover$^x$ & ineff & cover$^x$ & ineff& cover$^x$ & ineff\\\midrule
GAE&$0.9514\std{0.0144} $ 
&${3.3652}\std{0.1312}$
&$0.9517\std{0.0141}$
&$3.5878\std{0.2107}$
&$0.9578\std{0.0420}$
&$4.0504\std{1.2916}$
&$0.9524\std{0.0333}$
&$3.3292\std{0.5866}$\\
DiGAE&$0.9205\std{0.0498} $ 
&${3.3135}\std{0.1172}$
&$0.9223\std{0.0469}$
&$3.3872\std{0.1260}$
&$0.9250\std{0.0479} $ 
&$3.4241\std{0.1271}$
&$0.9089\std{0.0611}$
&$3.6158\std{0.2348}$\\
LGNN&$0.9284\std{0.0296}$ 
&${3.4362}\std{0.1029}$
&$0.9305\std{0.0258}$
&$3.4844\std{0.1233}$
&$0.9290\std{0.0284}$ 
&$3.6514\std{0.1050}$
&$0.9379\std{0.0261}$
&$4.0805\std{0.5445}$\\ \midrule  
Average & 0.9334 & 3.3716 & 0.9348 & 3.4865 & 0.9373 & 3.7086 & 0.9331 & 3.6752\\
\midrule
Score Method-CQR-cluster  & cover$^x$  & ineff & cover$^x$ & ineff & cover$^x$ & ineff& cover$^x$ & ineff\\\midrule
GAE&$0.9519
\std{0.0318}$
&$3.3721
\std{0.021}$
&$0.9532
\std{0.028}$
&$3.4862
\std{0.035}$
&$0.9557
\std{0.024}$
&$3.7083
\std{0.041}$
&$0.9541
\std{0.032}$
&$3.6749
\std{0.019}$\\ 
DiGAE&$0.9412
\std{0.025}$
&$3.3645
\std{0.018}$
&$0.9428
\std{0.031}$
&$3.4821
\std{0.027}$
&$0.9443
\std{0.029}$
&$3.7058
\std{0.033}$
&$0.9437
\std{0.026}$
&$3.6724
\std{0.022}$\\ 
LGNN&$0.9315
\std{0.037}$
&$3.3582
\std{0.015}$
&$0.9332
\std{0.034}$
&$3.4789
\std{0.029}$
&$0.9351
\std{0.031}$
&$3.7023
\std{0.036}$
&$0.9345
\std{0.028}$
&$3.6698
\std{0.024}$\\  
\midrule  
Average & 0.9415&  3.3649&  0.9424&  3.4824&  0.9450&  3.7055&  0.9438&  3.6720\\
\midrule
Score Method-CQR-RR  & cover$^x$  & ineff & cover$^x$ & ineff & cover$^x$ & ineff& cover$^x$ & ineff\\\midrule
GAE&$0.9482
\std{0.019}$
&$3.3018
\std{0.017}$
&$0.9497
\std{0.021}$
&$3.3976
\std{0.023}$
&$0.9513
\std{0.016}$
&$3.4241
\std{0.025}$
&$0.9508
\std{0.018}$
&$3.5372
\std{0.020}$\\ 
DiGAE&$0.9395
\std{0.026}$
&$3.2954
\std{0.019}$
&$0.9411
\std{0.028}$
&$3.3921
\std{0.024}$
&$0.9428
\std{0.022}$
&$3.4207
\std{0.027}$
&$0.9432
\std{0.025}$
&$3.5346
\std{0.021}$\\ 
LGNN&$0.9316
\std{0.035}$
&$3.2893
\std{0.014}$
&$0.9335
\std{0.032}$
&$3.3875
\std{0.026}$
&$0.9357
\std{0.029}$
&$3.4174
\std{0.028}$
&$0.9364
\std{0.027}$
&$3.5319
\std{0.023}$\\  
\midrule  
Average & 0.9442&  3.2945&  0.9414&  3.3920&  0.9433&  3.4207&  0.9435&  3.5346\\
\midrule
Score Method-CQR-RR-Cluster  & cover$^x$  & ineff & cover$^x$ & ineff & cover$^x$ & ineff& cover$^x$ & ineff\\\midrule
GAE&$0.9578\std{0.0134} $ 
&${3.1297}\std{0.1401}$
&$0.9578\std{0.0189}$
&$3.0985\std{0.1478}$
&$0.9527\std{0.0123}$ 
&$3.1614\std{0.1622}$
&$0.9520\std{0.0145}$
&$ \textbf{2.8927}\std{0.1223}$\\
DiGAE&$0.9513\std{0.0415} $ 
&$\textbf{3.0262}\std{0.1412}$
&$0.9501\std{0.0312}$
&$\textbf{2.8976}\std{0.1393} $
&$0.9507\std{0.0456}$
&$2.9347\std{0.1139}$
&$0.9442\std{0.0735}$ 
&$3.0321\std{0.2134}$\\
LGNN&$0.9438\std{0.0396}$
&${3.3562}\std{0.0355}$
&$0.9473\std{0.0423}$ 
&$3.1422\std{0.0423}$
&$0.9497\std{0.0323}$
&$\textbf{2.9913}\std{0.0732}$
&$0.9507\std{0.0324}$
&$3.5195\std{0.1231}$\\
\midrule

Average & \underline{\textbf{0.9510}} & \underline{\textbf{3.1707}} & \underline{\textbf{0.9517}} & \underline{\textbf{3.0461}} & \underline{\textbf{0.9510}} & \underline{\textbf{3.0291}} & \underline{\textbf{0.9490}} & \underline{\textbf{3.1481}}\\
\bottomrule

\end{tabular}    
\end{adjustbox}
\end{table*}

\subsection{RR-GNN on Node Regression}\label{sec: problem2}
We apply RR-GNN for the node regression task to predict a continuous target variable $y_i$ associated with each node $i$ in a graph. 
Firstly, we train a traditional GNN model (Conformal GNN) for the node regression task using the training set. The GNN model learns a function $f: \gG \rightarrow \displaystyle \R^n$, where $\gG$ is the input graph and $f(\gG)_i$ represents the predicted target variable for node $i$.  Node regression minimizes the distance between the direct output and labels. 
\begin{align}
\hat{\vy}  &= f_{\text{GNN}}(\mathbf{X}, \mathbf{A
}) ~\label{enbedding_equation} \\
\mathcal{L} &= |\vy - \hat{\vy}|^2
\end{align}
Here, we use the GNN-based model, GAE, to deal with several cases, which generates the embedding using an encoder function and generate node regression using the decoder function.
For the following steps, a Residual GNN predicts the residual of the node labels, generating the weight for the non-conformity measure when computing prediction sets. The details of the algorithm are shown in appendices Algorithm 2. %~\ref{alg: nodd regression}.

\subsection{RR-GNN on Node Classification}\label{sec: problem3}
The node classification problem is a fundamental task in graph-based machine learning, where the goal is to predict a discrete label or class for each node in a given graph.

We first trained the Conformal GNN model, a function $f: \displaystyle \gG \rightarrow \displaystyle \{0,1,...,K\}^n$, that maps the node features to the corresponding labels with K classes.
Compared with regression, node classification uses binary cross-entropy loss as the loss function:
\begin{align}
\hat{\vy} &= f_{\text{GNN}}(\mX,\mA) \\
\mathcal{L} &= -\frac{1}{N}\sum_{i=1}^{N}\sum_{k=1}^{K}y_{i,k}\log(\hat{y}_{i,k})
\end{align}
We then generate the residual of the predicted value and the actual label as the label for the Residual GNN. We do a softmax operation to get a vector, representing the probability of the node belonging to each class:
\begin{align}
\hat{\vl} &= \text{softmax}(\hat{\vy})
\end{align}
The residual $\hat{\vr}$ is obtained by:
\begin{align}
\hat{\vr} &= \hat{\vl}- \vy
\end{align}
Where $\vy$ is the one-hot label of which class the node belongs to. 
% ~\ref{alg: nodd classification} 
In Algorithm 4 appendices material, we set a small positive real number $\epsilon$ (1e-9) to avoid the denominator equal to 0.  In addition, we need the differentiable quantile method in equation (21). Since the non-conformity score is usually differentiable, it only requires differentiable quantile calculation where there are well-established methods available \cite{chernozhukov2010quantile,blondel2020fast}.
\begin{table*}[ht]
\caption{Results of RR-GNN on Node Regression Datasets}
\label{tab:eff_all_models12}
\centering
\begin{adjustbox}{width=0.85\textwidth}
\begin{tabular}{|l|c|c|c|c|c|c|c|c|}
\toprule
Dataset& \multicolumn{2}{c|}{GraphSAGE} & \multicolumn{2}{c|}{SGC}  & \multicolumn{2}{c|}{GCN} & \multicolumn{2}{c|}{GATS} \\ \cmidrule{1-9}
 Metrics & cover$^x$  & ineff & cover$^x$ & ineff & cover$^x$ & ineff& cover$^x$ & ineff\\\midrule
Anaheim: CF-GNN&$0.9520\std{0.0669} $ 
&$\textbf{1.9231}\std{0.0483}$
&$0.9559\std{0.0617}$
&$2.2031\std{0.0241}$
&$0.9519\std{0.0531} $ 
&$2.3782\std{0.0533}$
&$0.9523\std{0.0302}$
&$2.1499\std{0.0463}$\\
Anaheim: Cluster-GNN&$0.9532\std{0.042}$
&$1.8954\std{0.037}$
&$0.9561\std{0.035}$
&$2.1423\std{0.031}$
&$0.9528\std{0.041}$
&$2.2451\std{0.029}$
&$0.9541\std{0.028}$
&$2.0321\std{0.025}$\\
Anaheim: RR-GAE&$0.9539\std{0.038}$
&$1.8732\std{0.032}$
&$\textbf{0.9567}\std{0.031}$
&$2.0987\std{0.028}$
&$0.9532\std{0.036}$
&$2.1934\std{0.026}$
&$0.9563\std{0.024}$
&$1.9623\std{0.022}$\\
Anaheim: Clsuter-RR-GAE&$\textbf{0.9543}\std{0.0320} $ 
&${1.9647}\std{0.0197}$
&$0.9577\std{0.0657}$
&$\textbf{2.0188}\std{0.0246}$
&$\textbf{0.9585}\std{0.0413}$ 
&$\textbf{2.2179}\std{0.0254}$
&$\textbf{0.9638}\std{0.0302}$
&$\underline{\textbf{1.8996}}\std{0.0249}$\\
\midrule
Chicago: CF-GNN&$0.9448\std{0.0519} $ 
&${2.3426}\std{0.0384}$
&$0.9486\std{0.0247}$
&$1.0423\std{0.0372}$
&$0.9505\std{0.0447}$
&$2.0456\std{0.0443}$
&$0.9508\std{0.0569}$
&$1.1396\std{0.0686}$\\
Chicago: Cluster-GNN&$0.9461\std{0.039}$
&$2.2894\std{0.034}$
&$0.9492\std{0.031}$
&$1.1895\std{0.029}$
&$0.9513\std{0.037}$
&$1.8742\std{0.031}$
&$0.9516\std{0.042}$
&$1.1254\std{0.045}$\\
Chicago: RR-GAE&$0.9472\std{0.035}$
&$2.2673\std{0.029}$
&$\textbf{0.9498}\std{0.028}$
&$1.2567\std{0.026}$
&$0.9519\std{0.033}$
&$1.6923\std{0.027}$
&$0.9519\std{0.038}$
&$1.1489\std{0.039}$\\
Chicago: Cluster-RR-GAE&$\textbf{0.9476}\std{0.0426} $ 
&$\textbf{2.2291}\std{0.0325}$
&$0.9546\std{0.0328}$
&$\textbf{1.2012}\std{0.0250}$
&$\textbf{0.9538}\std{0.0356}$ 
&$\underline{\textbf{1.5769}}\std{0.0252}$
&$\textbf{0.9540}\std{0.0362}$  
&$1.1283\std{0.0256}$\\ \midrule  
Education: CF-GNN&$0.9501\std{0.0242} $ 
&${2.3808}\std{0.0427}$
&$0.9500\std{0.0285}$
&$2.4892\std{0.0351}$
&$0.9483\std{0.0408}$
&$2.4380\std{0.0452}$
&$0.9502\std{0.0392}$
&$2.4209\std{0.0376}$\\
Education: Cluster-GNN&$0.9513\std{0.031}$
&$2.3145\std{0.038}$
&$0.9517\std{0.033}$
&$2.3721\std{0.032}$
&$0.9496\std{0.035}$
&$2.2894\std{0.034}$
&$0.9518\std{0.036}$
&$2.3256\std{0.033}$\\
Education: RR-GAE&$0.9529\std{0.029}$
&$2.1932\std{0.027}$
&$0.9534\std{0.030}$
&$2.1478\std{0.028}$
&$0.9508\std{0.032}$
&$2.0321\std{0.029}$
&$0.9532\std{0.031}$
&$2.1423\std{0.030}$\\
Education: Cluster-RR-GAE&$\textbf{0.9599}\std{0.0417} $ 
&$\textbf{2.0573}\std{0.0280}$
&$\textbf{0.9586}\std{0.0225}$
&$\textbf{2.0445}\std{0.0239}$
&$\textbf{0.9580}\std{0.0333}$ 
&$\textbf{1.8731}\std{0.0260}$
&$\textbf{0.9594}\std{0.0386}$  
&$\underline{\textbf{1.9075}}\std{0.0221}$\\ \midrule
Election: CF-GNN&$0.9498\std{0.0211} $ 
&${0.9268}\std{0.0429}$
&$0.9495\std{0.0215}$
&$0.9279\std{0.0302}$
&$0.9506\std{0.0473}$
&$0.9009\std{0.0282}$
&$0.9488\std{0.0363}$
&$0.9136\std{0.0681}$\\
Election: Cluster-GNN&$0.9503\std{0.028}$
&$0.9152\std{0.038}$
&$0.9501\std{0.027}$
&$0.9124\std{0.035}$
&$0.9512\std{0.041}$
&$0.8723\std{0.031}$
&$0.9496\std{0.033}$
&$0.8945\std{0.042}$\\
Election: RR-GAE&$0.9509\std{0.025}$
&$0.9037\std{0.029}$
&$0.9523\std{0.024}$
&$0.8956\std{0.028}$
&$0.9518\std{0.036}$
&$0.8234\std{0.026}$
&$0.9514\std{0.030}$
&$0.8562\std{0.035}$\\
Election: Cluster-RR-GAE&$\textbf{0.9558}\std{0.0215}$ 
&$\textbf{0.9213}\std{0.0279}$
&$\textbf{0.9567}\std{0.0242}$
&$\textbf{0.9487}\std{0.0259}$
&$\textbf{0.9510}\std{0.0432}$ 
&$\textbf{0.9343}\std{0.0341}$
&$\textbf{0.9567}\std{0.0317}$  
&$\underline{\textbf{0.6698}}\std{0.0201}$\\ \midrule

\end{tabular}    
    \end{adjustbox}    
\end{table*}
\section{Results} \label{sec: r}
\subsection{Empirical analysis}\label{sec: empirical}
In this section, we showcase the application of the proposed RR-GNN on 15 datasets for edge weight prediction, node regression, and node classification problems. We conduct a comparative
analysis of the performance of RR-GNN and four competitors based on two metrics. For the data split, 30\% of the data was designated for training, 30\% for validation, 20\% for testing, and the remaining 20\% for calibration.

% \subsection{Transportation Network Snapshot}
\noindent
{\bf Datasets:} To evaluate the effectiveness of RR-GNN algorithm, we conduct experiments on four categories of benchmark graph datasets: 1) traffic datasets, 2) citation connection datasets, 3) social network datasets, and 4) additional datasets.  

We apply RR-GNN on the traffic network and traffic flow data from 
Chicago and Anaheim to predict each node's edge weight and traffic volume \cite{bar2021transportation}. Chicago dataset consists of 541 nodes representing road junctions and 2150 edges representing road segments with directions, while the Anaheim dataset consists of 413 nodes and 858 edges.
In this context, each node is characterized by a two-dimensional feature $\mX_{i,:}\in \displaystyle \R^{2}$ representing its coordinates, and each edge is associated with a weight that signifies the traffic volume passing through the corresponding road segment. We collect three widely used citation network datasets for the citation datasets: Cora, PubMed, and CiteSeer. We apply RR-GNN to paper classification and citation prediction. Social network datasets like Twitch, CS, and Physics have become increasingly important resources for graph machine-learning research.

\begin{figure}[ht]
\centerline{\includegraphics[height=4cm,width=0.45\textwidth,clip=]{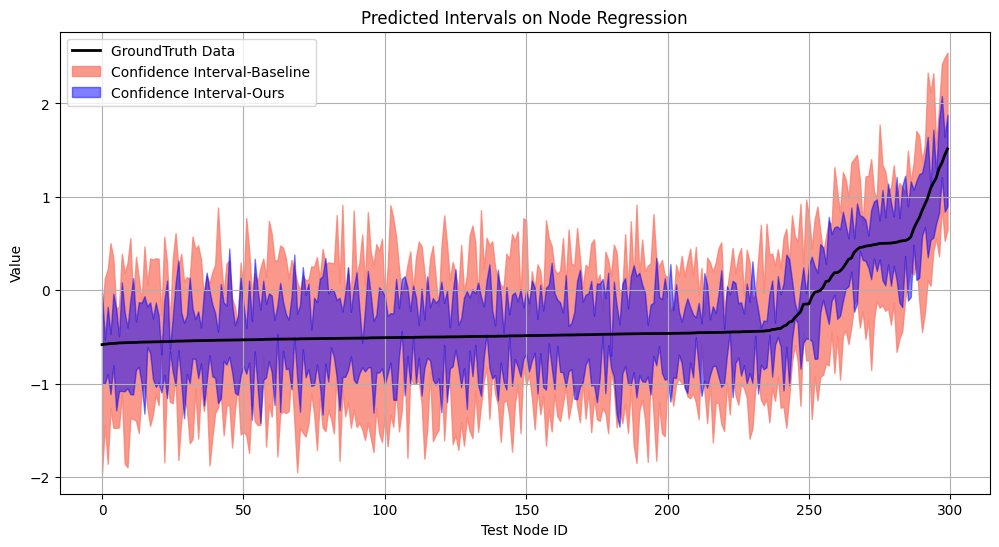}}

\caption{The prediction interval of node regression generated by CF-GNN and RR-GAE. The x-axis represents the node, which is sorted by the label. The y-axis represents the prediction intervals of nodes. The error rate $\alpha$ is 0.05. Blue and red represent the results of RR-GAE and CFNN-GAE, respectively. }

\label{fig: comparisonnr}
\end{figure}

\begin{figure}[ht]
\centerline{\includegraphics[height=5cm,width=0.5\textwidth,clip=]{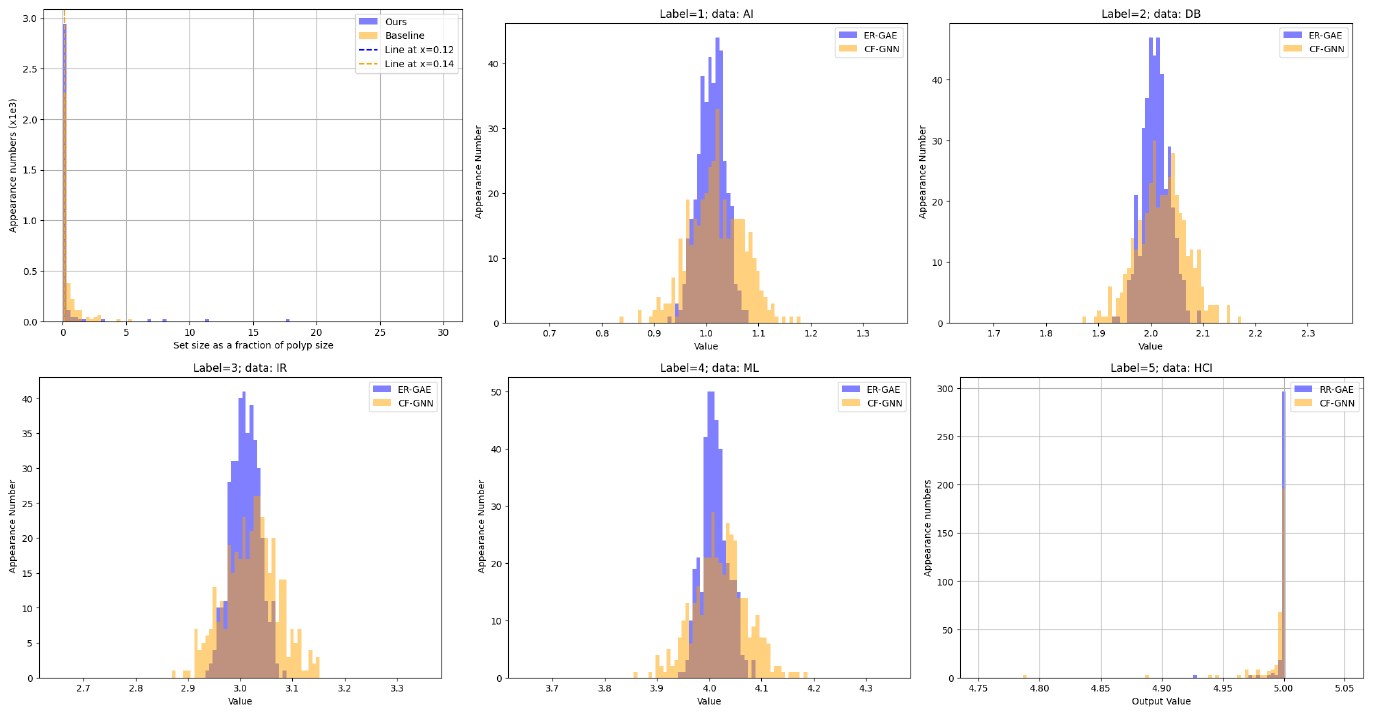}}

\caption{The histogram of predicted values of the node classification task on dataset MedPub. The error rate $\alpha$ is 0.05. Each sub-figure represents one classification from 0 to 5. The x-axis is the predicted value from the model. The y-axis is the frequency corresponding to the predicted value. Blue and yellow represent the results of RR-GAE and CF-GNN, respectively. }
\label{fig: comparisonnc}
\end{figure}

\textbf{Metrics:} We use inefficiency (ineff) and weighted symmetric calibration (WSC) as evaluation metrics (details of ineff and WSC can be accessed in appendices material). Lower ineff and higher WSC indicate better performance. To generate prediction intervals, we independently sample 1000 vectors $v$ from the unit sphere in $\displaystyle \R^{2m}$ space. The parameters $a, b, \delta$ are fine-tuned via grid search. Additionally, 25\% of test data is utilized to estimate optimal $v, a, b, \delta$ values. The conditional coverage is then calculated on the remaining 75\% of test data.

{\bf Models and baselines: }
There are three basic GNN models for the model: 1) GAE (Section \ref{subsec: GAE}), 2) line graph neural network (LGNN \cite{cai2021line})  and 3) a directed variant of GAE, called DiGAE.  We use two nonconformity score, including 1) CP \cite{huang2024uncertainty} and 2) conformal quantile regression (CQR). For the encoder in the basic GNN models, we choose 4 different structures: 1) GraphSAGE \cite{hamilton2017inductive}, 2) SAGEConv \cite{morris2019weisfeiler}, 3) GCN \cite{kipf2016semi}, and 4) GAT \cite{velivckovic2017graph}. We name the model that combines conformal prediction (CP, as described in work \cite{huang2024uncertainty}) with graph autoencoder (GAE, Section \ref{subsec: GAE}) as CP-GAE. Similarly, we name the model with the nonconformity score and the GNN models. For example, the models use CQR, and GAE is referred to as CQR-GAE.  As for node classification task, three models are added: HAN, Ensemble TS and CaGCN which are latest models baseline on this task. 
 Additionally, if the residual reweighting approach is performed based on the CQR approach,  we also add RR to the name of the models. For example, the residual reweighted CQR-GAE is referred to as CQR-RR-GAE. The baseline models are the above models without residual reweight.

We use four popular graph neural network (GNN) model structures for encoder and decoder - GCN \cite{kipf2016semi}, GraphConv \cite{morris2019weisfeiler}, GAT \cite{velivckovic2017graph}, and GraphSAGE \cite{hamilton2017inductive} - as the base graph convolution layers for both the CP and CQR based models.

{\bf Experiment result:} 
As for the task of edge weight prediction,  we ran the experiment 10 times and split the data into training, validation, and the combined calibration and test sets for each dataset and model. We conduct 100 random splits of calibration and testing edges to perform the baseline model and RR-GNN and evaluate the empirical coverage. Our method achieved 6.15\% to 28.87\% reduce on inefficiency interval length shown in table 1 and 2 on the task of edge weight prediction. In addition, we conducted a paired t-testing on the mean inefficiency values for our method (RR) compared to CP and CQR, taking the meaning value of each (12 values in total in Table 1) after repeating the experiments 10 times. The p-values are 0.00035 and 0.00024. Coverage is defined as the probability that the ground truth value lies within the predicted confidence interval. Our method allows control of the coverage through the parameter $\alpha$, where the expected coverage is $1-\alpha$. In our manuscript, we set $\alpha = 0.05$, corresponding to a target coverage of 0.95. As shown in Tables 1-3 of the manuscript, the empirical coverages achieved by our method are close to 0.95, indicating that the coverage is well controlled and aligns with the expected theoretical value. 

\begin{figure}[ht]
\centerline{\includegraphics[height=5cm,width=0.5\textwidth,clip=]{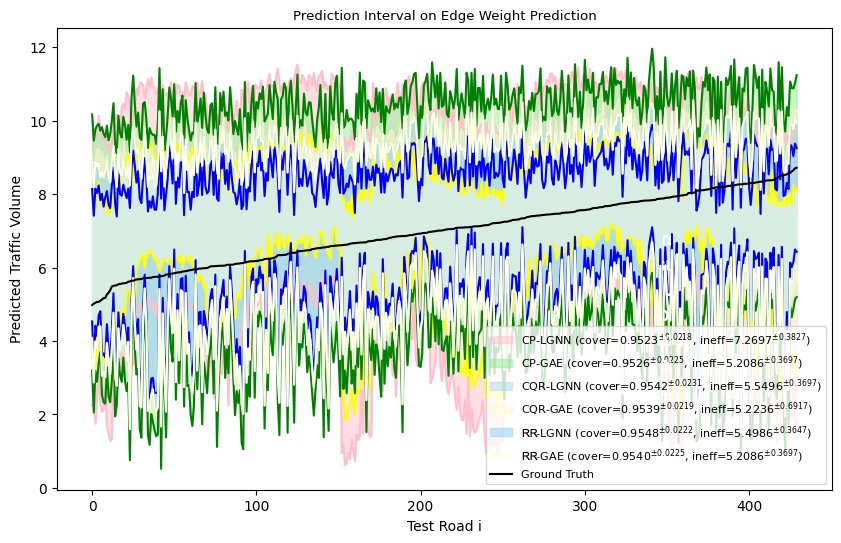}}
\caption{The graph shows the traffic volume prediction intervals generated on the Chicago traffic dataset. All methods set their error rate $\alpha$ at 0.05. The x-axis represents individual roads sorted by their actual/ground truth traffic volumes. The y-axis represents the predicted intervals. Different colors distinguish the results from different prediction methods. }
\label{fig: comparisonwep}
\end{figure}

We conducted a conditional coverage Equation 37 in appendices material) of RR-GNN and baseline methods on the Chicago and Anaheim traffic dataset for the edge weight prediction task.
The results presented in Table~\ref{tab:eff_all_models1gsmg} show that the overall RR-GNN models outperform others in terms of inefficiency (as defined in Equation 36 in appendices material) and conditional coverage  (Equation 37 in appendices material). This indicates that the RR variants can strike a better balance between capturing the uncertainty in the predictions and maintaining a high level of accuracy. We also find that GAE and LGNN outperform DiGAE, highlighting the efficacy of the autoencoder approach in weight prediction from Table 1. 
We showcase the prediction interval produced by model of LGNN with RR(CQR) in Figure~\ref{fig: comparisonwep}.  Furthermore, Figure~\ref{fig: comparisonwep} also illustrates the adaptability of the RR models by generating the smallest prediction intervals of varying sizes, which aligns with the data characteristics.

For the node regression task, we apply the models to 7 different datasets: 1) Anaheim traffic dataset, 2) Chicago traffic dataset, 3) Education dataset, 4) Income dataset, 5) Unemploy dataset, and 7) Twitch dataset. Table 4 in appendices material shows that our method outperforms the baseline model, CF-GNN \cite{huang2024uncertainty}, both on WSC score (coverage) and inefficiency on 7 datasets.  Besides, the visualization result in Figure~\ref{fig: comparisonnr} shows that we have a smaller interval size than that from CF-GNN. Similarly, for the task-node classification, we compared RR-GNN and CF-GNN on 8 datasets: 1) Cora, 2) DBLP, 3) CiteSeer, 4) PubMed, 5) Computer, 6) Photo, 7) CS and 8) Physics. Number and visual results can be seen in Table 4 and Figure~\ref{fig: comparisonnc}. Our model achieves better results in both accuracy and inefficiency. It is worth to mention that we employ categorical cross-entropy as the loss function for multi-class classification. The loss function is given by:
\begin{equation}
\mathcal{L} = -\frac{1}{N}\sum_{i=1}^{N}\sum_{k=1}^{K} y_{i,k} \log(\hat{y}_{i,k}),
\end{equation}
where $\hat{y}_{i,k}$ represents the predicted probability of node $i$ belonging to class $k$, and $\sum_{k=1}^{K} \hat{y}_{i,k} = 1$.

In summary, leveraging the RR-based models can generate prediction intervals that are both efficient and well-calibrated, making them a more suitable choice for different network-based tasks, especially real-world transportation applications where accurate and reliable predictions are crucial for informed decision-making.

Furthermore, we compared the predicted residual between CF-GNN and RR-GNN. Figure 7 in the appendices material shows the predicted residual of road's traffic volume for Chicago and Anathm. The residual value of CF-GNN is higher than that of RR-GAE. Figure 6 in appendices material also shows the residual of the U.S.A election result, where we can see that the global residual/difference between the model output and ground truth from RR-GAE is much lower than these baselines.

\begin{table*}[ht]
\caption{Results of Ours (RR-GNN) on Node Classification Datasets}
\label{tab:eff_all_models1}
\centering
\begin{adjustbox}{width=0.85\textwidth}
\begin{tabular}{|l|c|c|c|c|c|c|c|c|}
\toprule
Dataset& \multicolumn{2}{c|}{HAN} & \multicolumn{2}{c|}{SGC}  & \multicolumn{2}{c|}{CaGCN} & \multicolumn{2}{c|}{GATS} \\ \cmidrule{1-9}
Dataset  & cover$^x$  & ineff & cover$^x$ & ineff & cover$^x$ & ineff& cover$^x$ & ineff\\\midrule
Cora: CF-GNN&$0.9456\std{0.0569} $ 
&${1.6284}\std{0.0483}$
&$0.9461\std{0.0603}$
&$1.6633\std{0.0441}$
&$0.9473\std{0.0556} $ 
&$1.6344\std{0.0418}$
&$0.9464\std{0.0702}$
&$1.6278\std{0.0334}$\\
Cora: Cluster-GAE&$0.9458\std{0.0532} $ 
&${1.61201}\std{0.0431}$
&$0.9459\std{0.0612}$
&$1.6537\std{0.0432}$
&$0.9385\std{0.0529}$ 
&$1.6188\std{0.0328}$
&$0.9482\std{0.0453}$
&$1.6013\std{0.0313}$\\
Cora: RR-GAE&$0.9460\std{0.0542}$ 
&$1.6100\std{0.0415}$
&$0.9462\std{0.0581}$
&$1.6297\std{0.0428}$
&$0.9432\std{0.0573}$ 
&$1.6251\std{0.0367}$
&$0.9475\std{0.0624}$
&$1.6146\std{0.0351}$\\
Cora: Cluster-RR-GAE&$\textbf{0.9478}\std{0.0523} $ 
&$\textbf{1.5896}\std{0.0354}$
&$\textbf{0.9490}\std{0.0643}$
&$\textbf{1.5907}\std{0.0432}$
&$\textbf{0.9465}\std{0.0759} $ 
&$\textbf{1.6175}\std{0.0354}$
&$\textbf{0.9508}\std{0.0554}$
&$\textbf{1.6114}\std{0.0287}$\\ \midrule
DBLP: CF-GNN&$0.9501\std{0.0523} $ 
&${1.5723}\std{0.0683}$
&$\textbf{0.9451}\std{0.0617}$
&$1.5274\std{0.0416}$
&$0.9473\std{0.0596} $ 
&$1.5644\std{0.0733}$
&$0.9467\std{0.0717}$
&$1.5729\std{0.0463}$\\
DBLP: Cluster-GAE&$0.9497\std{0.0512}$ 
&$1.5489\std{0.0492}$
&$0.9457\std{0.0583}$
&$1.4873\std{0.0449}$
&$0.9452\std{0.0684}$ 
&$1.5569\std{0.0317}$
&$0.9479\std{0.0673}$
&$1.5814\std{0.0376}$\\
DBLP: RR-GAE&$0.9499\std{0.0531}$ 
&$1.5351\std{0.0473}$
&$0.9462\std{0.0528}$
&$1.4286\std{0.0541}$
&$0.9458\std{0.0702}$ 
&$1.5512\std{0.0295}$
&$0.9485\std{0.0589}$
&$1.5725\std{0.0349}$\\
DBLP: Cluster-RR-GAE&$\textbf{0.9518}\std{0.0509} $ 
&$\textbf{1.5467}\std{0.0427}$
&$0.9503\std{0.0428}$
&$\textbf{1.3563}\std{0.0626}$
&$\textbf{0.9484}\std{0.0624} $ 
&$\textbf{1.5371}\std{0.0248}$
&$\textbf{0.9505}\std{0.0469}$
&$\textbf{1.5570}\std{0.0356}$\\ \midrule
CiteSeer: CF-GNN&$0.9528\std{0.0203} $ 
&${1.1680}\std{0.0439}$
&$0.9525\std{0.0257}$
&$\textbf{1.1827}\std{0.0552}$
&$0.9496\std{0.0392}$
&$1.2310\std{0.0332}$
&$0.9508\std{0.0309}$
&$1.2396\std{0.0416}$\\
CiteSeer: Cluster-GAE&$0.9532\std{0.0218}$ 
&$1.1653\std{0.0427}$
&$0.9561\std{0.0274}$
&$1.1854\std{0.0483}$
&$0.9507\std{0.0365}$ 
&$1.2237\std{0.0311}$
&$0.9523\std{0.0332}$
&$1.2298\std{0.0384}$\\
CiteSeer: RR-GAE&$0.9538\std{0.0853}$ 
&$1.1621\std{0.0552}$
&$0.9579\std{0.0536}$
&$1.1782\std{0.0415}$
&$0.9512\std{0.0358}$ 
&$1.2189\std{0.0276}$
&$0.9535\std{0.0447}$
&$1.2085\std{0.0361}$\\
CiteSeer: Cluster-RR-GAE&$\textbf{0.9556}\std{0.0918} $ 
&$\textbf{1.1539}\std{0.0615}$
&$\textbf{0.9598}\std{0.0561}$
&$1.1678\std{0.0372}$
&$\textbf{0.9526}\std{0.0363}$ 
&$\textbf{1.2016}\std{0.0289}$
&$\textbf{0.9562}\std{0.0428}$  
&$\textbf{1.1408}\std{0.0361}$\\ \midrule  
PubMed: CF-GNN&$0.9502\std{0.0207} $ 
&${1.4680}\std{0.0361}$
&$0.9508\std{0.0276}$
&$1.4272\std{0.0325}$
&$0.9516\std{0.0458}$
&$1.5310\std{0.0514}$
&$0.9512\std{0.0434}$
&$1.4396\std{0.0485}$\\
PubMed: Cluster-GAE&$0.9507\std{0.0352}$ 
&$1.3985\std{0.0374}$
&$0.9513\std{0.0419}$
&$1.4083\std{0.0341}$
&$0.9519\std{0.0462}$ 
&$1.4521\std{0.0483}$
&$0.9514\std{0.0427}$
&$1.4198\std{0.0491}$\\
PubMed: RR-GAE&$0.9510\std{0.0386}$ 
&$1.3528\std{0.0357}$
&$0.9516\std{0.0453}$
&$1.3992\std{0.0328}$
&$0.9520\std{0.0469}$ 
&$1.3815\std{0.0301}$
&$0.9515\std{0.0432}$
&$1.4085\std{0.0503}$\\
PubMed: Cluster-RR-GAE&$\textbf{0.9526}\std{0.0483}$ 
&$\textbf{1.3275}\std{0.0392}$
&$\textbf{0.9520}\std{0.0482}$
&$\textbf{1.3897}\std{0.0339}$
&$\textbf{0.9521}\std{0.0473}$ 
&$\textbf{1.3732}\std{0.0296}$
&$\textbf{0.9515}\std{0.0419}$  
&$\textbf{1.3989}\std{0.0522}$\\ \midrule
Computers: CF-GNN&$0.9471\std{0.0276} $ 
&${3.3680}\std{0.3499}$
&$0.9492\std{0.0235}$
&$3.8272\std{0.0292}$
&$0.9457\std{0.0435}$
&$3.2310\std{0.0652}$
&$0.9478\std{0.0325}$
&$3.1396\std{0.0586}$\\
Computers: Cluster-GAE&$0.9476\std{0.0321}$ 
&$3.1523\std{0.3287}$
&$0.9490\std{0.0273}$
&$3.4821\std{0.0315}$
&$0.9461\std{0.0418}$ 
&$2.8945\std{0.0583}$
&$0.9479\std{0.0382}$
&$2.9634\std{0.0541}$\\
Computers: RR-GAE&$0.9481\std{0.0473}$ 
&$2.8937\std{0.0328}$
&$0.9493\std{0.0298}$
&$2.7324\std{0.0394}$
&$0.9464\std{0.0436}$ 
&$2.6745\std{0.0352}$
&$0.9479\std{0.0623}$
&$2.8033\std{0.0259}$\\
Computers: Cluster-RR-GAE&$\textbf{0.9503}\std{0.0553} $ 
&$\textbf{2.7423}\std{0.0258}$
&$\textbf{0.9509}\std{0.0315}$
&$\underline{\textbf{2.6343}}\std{0.0413}$
&$\textbf{0.9418}\std{0.0436}$ 
&$\textbf{2.5471}\std{0.0365}$
&$\textbf{0.9354}\std{0.0584}$  
&$\underline{\textbf{2.7739}}\std{0.0272}$\\ \midrule
Photo: CF-GNN&$0.9511\std{0.0275} $ 
&${3.2680}\std{0.0395}$
&$0.9515\std{0.0263}$
&$2.2276\std{0.0354}$
&$0.9486\std{0.0419}$
&$2.2010\std{0.0387}$
&$0.9509\std{0.0391}$
&$2.1986\std{0.0286}$\\
Photo: Cluster-GAE&$0.9523\std{0.0289}$ 
&$3.0125\std{0.0362}$
&$0.9517\std{0.0291}$
&$2.1224\std{0.0338}$
&$0.9491\std{0.0396}$ 
&$2.1076\std{0.0352}$
&$0.9510\std{0.0374}$
&$2.0059\std{0.0263}$\\
Photo: RR-GAE&$0.9527\std{0.0852}$ 
&$2.7843\std{0.0415}$
&$0.9518\std{0.0894}$
&$2.0451\std{0.0331}$
&$0.9495\std{0.0821}$ 
&$2.0128\std{0.0513}$
&$0.9511\std{0.0439}$
&$1.9015\std{0.0254}$\\
Photo: Cluster-RR-GAE&$\textbf{0.9554}\std{0.0723} $ 
&$\textbf{2.5474}\std{0.0456}$
&$\textbf{0.9534}\std{0.0913}$
&$\textbf{2.0026}\std{0.0316}$
&$\textbf{0.9504}\std{0.0342} $ 
&$\textbf{2.0003}\std{0.0370}$
&$\textbf{0.9498}\std{0.0512}$  
&$\textbf{1.7093}\std{0.0234}$\\ \midrule
CS: CF-GNN&$0.9438\std{0.0224} $ 
&${1.8669}\std{0.0347}$
&$0.9435\std{0.0284}$
&$1.6272\std{0.0452}$
&$0.9476\std{0.0416}$
&$3.6310\std{0.0325}$
&$0.9478\std{0.0317}$
&$2.7396\std{0.0286}$\\
CS: Cluster-GAE&$0.9451\std{0.0253}$ 
&$1.8324\std{0.0332}$
&$0.9448\std{0.0316}$
&$1.6229\std{0.0428}$
&$0.9483\std{0.0387}$ 
&$3.1957\std{0.0301}$
&$0.9481\std{0.0293}$
&$2.5641\std{0.0269}$\\
CS: RR-GAE&$0.9472\std{0.0573}$ 
&$1.8453\std{0.0365}$
&$0.9461\std{0.0528}$
&$1.6205\std{0.0384}$
&$0.9435\std{0.0546}$ 
&$2.8932\std{0.0275}$
&$0.9483\std{0.0362}$
&$2.4785\std{0.0241}$\\
CS: Cluster-RR-GAE&$\textbf{0.9502}\std{0.0601} $ 
&$\textbf{1.8430}\std{0.0361}$
&$\textbf{0.9501}\std{0.0528}$
&$\textbf{1.6183}\std{0.0361}$
&$\textbf{0.9516}\std{0.0525} $ 
&$\underline{\textbf{2.5469}}\std{0.0227}$
&$\textbf{0.9485}\std{0.0329}$  
&$\textbf{2.3889}\std{0.0238}$\\ \midrule
Physics: CF-GNN&$0.9495\std{0.0243} $ 
&${1.2218}\std{0.0463}$
&$0.9507\std{0.0292}$
&$1.2430\std{0.0324}$
&$0.9489\std{0.0257} $ 
&$1.2005\std{0.0604}$
&$0.9505\std{0.0275}$
&$1.2243\std{0.0246}$\\
Physics: Cluster-GAE&$0.9498\std{0.0267}$ 
&$1.2205\std{0.0428}$
&$0.9510\std{0.0319}$
&$1.2418\std{0.0346}$
&$0.9491\std{0.0283}$ 
&$1.2069\std{0.0551}$
&$0.9506\std{0.0298}$
&$1.2231\std{0.0239}$\\
Physics: RR-GAE&$0.9501\std{0.0573}$ 
&$1.2198\std{0.0283}$
&$0.9512\std{0.0501}$
&$1.2412\std{0.0385}$
&$0.9493\std{0.0326}$ 
&$1.2145\std{0.0423}$
&$0.9507\std{0.0442}$
&$1.2298\std{0.0249}$\\
Physics: Cluster-RR-GAE&$\textbf{0.9518}\std{0.0511}$ 
&$\textbf{1.2050}\std{0.0223}$
&$\textbf{0.9528}\std{0.0542}$
&$\textbf{1.2279}\std{0.0419}$
&$\textbf{0.9508}\std{0.0334}$ 
&$\textbf{1.1998}\std{0.0438}$
&$\textbf{0.9522}\std{0.0493}$
&$\textbf{1.2187}\std{0.0238}$\\ \bottomrule
\end{tabular}    
\end{adjustbox}    
\end{table*}

\subsection{Ablation Study}
To address the computational cost concerns, we tested the time and memory usage on random graphs and analyzed the time complexity of the algorithm.

Specifically, we tested three types of random graphs: Erdős-Rényi~\cite{erdos1961evolution}, Barabási-Albert~\cite{lei2018distribution}, and Watts-Strogatz small-world model~\cite{romano2019conformalized}. For each type, we generated graphs with node counts of 100, 10000, 50000, and 100000, respectively. Each node had 128 features, and the models were trained for 100 epochs. We present results using GAE (Graph Autoencoder) and DiGAE, combined with three different GNN (Graph Neural Network) encoders: GraphSAGE, GCN (Graph Convolutional Network), and GAT (Graph Attention Network) shown in Table\ref{t-ews}. The used GPU space is increasing linearly along with the node size, while processing time increases at a rate lower than linear with respect to the size. Our findings indicate that the time and space requirements are within acceptable limits for these configurations.

Our method, RR-GNN, involves training two GNN models with a cross-training protocol. Subsequently, RR-GNN uses Louvain clustering to generate interval predictions. The time complexity of the GNN component is $O(m)$, where $m$ represents the number of edges~\cite{wu2020comprehensive}. The Louvain clustering algorithm has a time complexity of $O(m \log m)$ ~\cite{kumar2016comparing}. For the cross-training protocol, we set a fixed number of training iterations. Therefore, the overall time complexity of RR-GNN can be approximated as $O(m \log m)$. Besides, in order to evaluate our method with the best current methods, we list the results in the appendices material. We test Graphormer and GT in tables 6 and 7 in the appendices material. We can see that our RR operation can help improve the performance of Graphormer and GT. In addition, we compare SAN, which is a transformer-based method in Table 7 in the appendices section. The result shows that our RR-GCN is better than SAN overall.

\section{Conclusion}\label{sec: conclusion}
This paper introduces the Residual Reweighted Conformal Prediction Graph Neural Network (RR-GNN), which enhances graph neural networks (GNNs) by integrating conformal prediction (CP). While traditional GNNs yield point predictions, RR-GNN provides predictive regions reflecting varying confidence levels. Existing nonconformity measures often produce uniform-width regions, neglecting the differing prediction difficulties. RR-GNN overcomes this by employing a novel residual reweighting nonconformity measure that adjusts predictive region widths based on expected accuracy for each example. We validate RR-GNN's effectiveness on 15 datasets, including real-world datasets, like transportation and social networks, across tasks like edge weight prediction and node classification. RR-GNN consistently delivers tighter predictive regions, higher accuracy, and improved efficiency compared to standard GNN methods, advancing uncertainty-aware predictions in graph machine learning.

\section*{Acknowledgment}

This work was partially supported by Hong Kong RGC, City University of Hong Kong grants (Project No. 9610639 and 6000864), and Chengdu Municipal Office of Philosophy and Social Science grant 2024BS013.

\section{APPENDICES}

%\begin{figure*}[ht]
%\centerline{\includegraphics[height=5cm, width=\textwidth,clip=]{figures/figure1.pdf}}
%\caption{Schematic figure for transductive and inductive settings for edge weight prediction.
%Different colors indicate the availability of the nodes during the training or testing phases.
%Solid and dashed lines represent edges used for training and the predicted edge in the testing phases, respectively.
%(a) Transductive edge weight prediction performs both training and inference on the same graph.
%(b) Inductive edge weight prediction inference is performed on a new, unseen graph.}
%\label{fig: transductive}
%\end{figure*}
\subsection{Other Experiment Results}
In order to test our method's effcience on more datasets, we did 4 experiments. First is the extended test on node regression on other four datasets shown as~\ref{tab:eff_all_nr_rest}. The second result shows node classification results comparing with conformal baselines can be found in Table~\ref{tab:core_results}. Thirdly, in order to prove our method's generlization, we incorporated additional GNN models into our experiments using the Anaheim Data and Chicago Data datasets. It needs to be clarified that our method can be combined with all other GNN models (Graphormer [1], GT [2]).  Details are shown Table~\ref{tab:transformer_rr_comparison} ~\ref{tab:GT_rr_comparison}.
\begin{table*}[ht]
\caption{Results of RR-GNN on Node Regression Datasets}
\label{tab:eff_all_nr_rest}
\centering
\begin{adjustbox}{width=\textwidth}
\begin{tabular}{|l|c|c|c|c|c|c|c|c|}
\toprule
Dataset& \multicolumn{2}{c|}{GraphSAGE} & \multicolumn{2}{c|}{SGC}  & \multicolumn{2}{c|}{GCN} & \multicolumn{2}{c|}{GATS} \\ \cmidrule{1-9}
 Metrics & cover$^x$  & ineff & cover$^x$ & ineff & cover$^x$ & ineff& cover$^x$ & ineff\\\midrule
Income: CF-GNN&$0.9512\std{0.0264} $ 
&${2.7580}\std{0.0342}$
&$0.9504\std{0.0405}$
&$2.4892\std{0.0302}$
&$0.9511\std{0.0250}$
&$2.5272\std{0.0318}$
&$0.9508\std{0.0329}$
&$2.4396\std{0.0328}$\\
Income: Cluster-GNN&$0.9521\std{0.035}$
&$2.6723\std{0.041}$
&$0.9513\std{0.038}$
&$2.3721\std{0.037}$
&$0.9526\std{0.033}$
&$2.4189\std{0.036}$
&$0.9519\std{0.034}$
&$2.3254\std{0.035}$\\
Income: RR-GAE&$0.9538\std{0.032}$
&$2.5342\std{0.038}$
&$\textbf{0.9524}\std{0.036}$
&$2.1423\std{0.034}$
&$0.9539\std{0.031}$
&$2.1932\std{0.033}$
&$0.9527\std{0.033}$
&$2.1567\std{0.032}$\\
Income: Cluster-RR-GAE&$\textbf{0.9552}\std{0.0618}$&$\textbf{2.1003}\std{0.0492}$
&$0.9519\std{0.0513}$
&$\textbf{1.9616}\std{0.0358}$
&$\textbf{0.9566}\std{0.0501} $ 
&$\textbf{1.9203}\std{0.0354}$
&$\textbf{0.9545}\std{0.0347}$  
&$\underline{\textbf{1.8555}}\std{0.0423}$\\ \midrule
Unemploy: CF-GNN&$0.9526\std{0.0415}$ 
&${2.2298}\std{0.0523}$
&$0.9510\std{0.0320}$
&$2.4587\std{0.0491}$
&$0.9506\std{0.0294}$
&$2.5013\std{0.0326}$
&$0.9502\std{0.0354}$
&$2.4332\std{0.0376}$\\
Unemploy: Cluster-GNN&$0.9531\std{0.038}$
&$2.1932\std{0.045}$
&$0.9519\std{0.036}$
&$2.3256\std{0.042}$
&$0.9513\std{0.034}$
&$2.3721\std{0.038}$
&$0.9516\std{0.033}$
&$2.2894\std{0.039}$\\
Unemploy: RR-GAE&$0.9542\std{0.035}$
&$2.1423\std{0.039}$
&$0.9524\std{0.033}$
&$2.1932\std{0.036}$
&$0.9528\std{0.032}$
&$2.2567\std{0.035}$
&$0.9523\std{0.031}$
&$2.1567\std{0.034}$\\
Unemploy: Cluster-RR-GAE&$\textbf{0.9569}\std{0.0419}$ 
&$\textbf{2.0816}\std{0.0218}$
&$\textbf{0.9517}\std{0.0313}$
&$\textbf{2.0534}\std{0.0367}$
&$\textbf{0.9523}\std{0.0369}$ 
&$\textbf{2.0480}\std{0.0190}$
&$\textbf{0.9523}\std{0.0448}$  
&$\underline{\textbf{1.9503}}\std{0.0312}$\\ \midrule
Twitch: CF-GNN&$\textbf{0.9524}\std{0.0443} $ 
&$2.6634\std{0.0365}$
&$0.9523\std{0.0392}$
&$2.6835\std{0.0394}$
&$0.9529\std{0.0257} $ 
&$2.5409\std{0.0404}$
&$0.9515\std{0.0275}$
&$2.6243\std{0.0460}$\\
Twitch: Cluster-GNN&$0.9531\std{0.039}$
&$2.5894\std{0.042}$
&$0.9528\std{0.037}$
&$2.5321\std{0.040}$
&$0.9534\std{0.034}$
&$2.4892\std{0.038}$
&$0.9523\std{0.033}$
&$2.4723\std{0.041}$\\
Twitch: RR-GAE&$0.9539\std{0.036}$
&$\textbf{2.4987}\std{0.039}$
&$\textbf{0.9532}\std{0.035}$
&$2.4567\std{0.037}$
&$0.9541\std{0.032}$
&$2.3721\std{0.036}$
&$0.9529\std{0.031}$
&$2.3256\std{0.038}$\\
Twitch: Cluster-RR-GAE&$0.9515\std{0.0367}$ 
&${5.0491}\std{0.0513}$
&$0.9541\std{0.0284}$
&$\underline{\textbf{2.1005}}\std{0.0189}$
&$\textbf{0.9571}\std{0.0219}$ 
&$\textbf{2.2398}\std{0.0225}$
&$\textbf{0.9535}\std{0.0280}$
&$\textbf{2.1353}\std{0.0262}$\\ \bottomrule

\end{tabular}    
    \end{adjustbox}    
\end{table*}

\begin{table*}[ht]
\caption{Node Classification Results with Conformal Baselines (Coverage $\uparrow$ / Inefficiency $\downarrow$)}
\label{tab:core_results}
\centering
\begin{adjustbox}{width=\textwidth}
\begin{tabular}{|l|c|c|c|c|c|c|c|c|}
\toprule
Dataset & \multicolumn{2}{c|}{CF-GNN [1]} & \multicolumn{2}{c|}{SAN} & \multicolumn{2}{c|}{RR-GNN (Ours)} & \multicolumn{2}{c|}{Cluster-RR-GNN (Ours)} \\ 
\cmidrule(lr){2-3} \cmidrule(lr){4-5} \cmidrule(lr){6-7} \cmidrule(lr){8-9}
Model & Cover & Ineff & Cover & Ineff & Cover & Ineff & Cover & Ineff \\
\midrule
\textbf{Cora} & & & & & & & & \\
GraphSAGE & $0.9456\std{0.0569}$ & $1.6284\std{0.0483}$ & $0.9476\std{0.0532}$ & $1.6825\std{0.0541}$ & $0.9460\std{0.0542}$ & $1.6100\std{0.0415}$ & $\textbf{0.9463}\std{0.0509}$ & $\textbf{1.6076}\std{0.0397}$ \\
SGC & $0.9461\std{0.0603}$ & $1.6633\std{0.04415}$ & $0.9482\std{0.05348}$ & $1.6956\std{0.0236}$ & $0.9462\std{0.0581}$ & $1.6297\std{0.0428}$ & $\textbf{0.9468}\std{0.0662}$ & $\textbf{1.6017}\std{0.0465}$ \\
GCN & $0.9473\std{0.0556}$ & $1.6344\std{0.0418}$ & $0.9426\std{0.0453}$ & $1.6520\std{0.0344}$ & $0.9432\std{0.0573}$ & $1.6251\std{0.0367}$ & $\textbf{0.9476}\std{0.0732}$ & $\textbf{1.6315}\std{0.0303}$ \\
GAT & $0.9464\std{0.0702}$ & $1.6278\std{0.0334}$ & $0.9473\std{0.065}$ & $1.6752\std{0.0364}$ & $0.9475\std{0.0624}$ & $1.6146\std{0.0351}$ & $\textbf{0.9491}\std{0.0539}$ & $\textbf{1.6254}\std{0.0396}$ \\
\midrule
\textbf{DBLP} & & & & & & & & \\
GraphSAGE & $0.9501\std{0.0523}$ & $1.5723\std{0.0683}$ & $0.9501\std{0.0420}$ & $\textbf{1.5524}\std{0.0637}$ & $0.9499\std{0.0531}$ & $1.5351\std{0.0473}$ & $\textbf{0.9503}\std{0.0510}$ & $1.5607\std{0.0487}$ \\
SGC & $\textbf{0.9451}\std{0.0617}$ & $1.5274\std{0.0416}$ & $0.9235\std{0.0526}$ & $1.4520\std{0.0345}$ & $0.9462\std{0.0528}$ & $1.4286\std{0.0541}$ & $0.9443\std{0.0462}$ & $\textbf{1.3921}\std{0.0624}$ \\
GCN & $\textbf{0.9473}\std{0.0596}$ & $1.5644\std{0.0733}$ & $0.9445\std{0.0565}$ & $1.5984\std{0.0743}$ & $0.9458\std{0.0702}$ & $1.5512\std{0.0295}$ & $0.9430\std{0.0713}$ & $\textbf{1.5491}\std{0.0278}$ \\
GAT & $0.9467\std{0.0717}$ & $1.5729\std{0.0463}$ & $0.9456\std{0.0624}$ & $1.5943\std{0.0425}$ & $0.9485\std{0.0589}$ & $1.5725\std{0.0349}$ & $\textbf{0.9491}\std{0.0539}$ & $\textbf{1.5720}\std{0.0322}$ \\
\midrule
\textbf{CiteSeer} & & & & & & & & \\
GraphSAGE & $0.9528\std{0.0203}$ & $1.1680\std{0.0439}$ & $0.9502\std{0.0164}$ & $1.2523\std{0.0416}$ & $0.9538\std{0.0853}$ & $1.1621\std{0.0552}$ & $\textbf{0.9540}\std{0.0926}$ & $\textbf{1.1679}\std{0.0605}$ \\
SGC & $0.9525\std{0.0257}$ & $1.1827\std{0.0552}$ & $0.9505\std{0.0645}$ & $1.2678\std{0.0532}$ & $0.9579\std{0.0536}$ & $1.1782\std{0.0415}$ & $\textbf{0.9594}\std{0.0582}$ & $\textbf{1.1898}\std{0.0399}$ \\
GCN & $0.9496\std{0.0392}$ & $1.2310\std{0.0332}$ & $0.9502\std{0.0536}$ & $1.3024\std{0.0324}$ & $0.9512\std{0.0358}$ & $1.2189\std{0.0276}$ & $0.9518\std{0.0373}$ & $\textbf{1.2153}\std{0.0290}$ \\
GAT & $0.9508\std{0.0309}$ & $1.2396\std{0.0416}$ & $0.9515\std{0.0251}$ & $1.3112\std{0.0123}$ & $0.9535\std{0.0447}$ & $1.2085\std{0.0361}$ & $\textbf{0.9548}\std{0.0491}$ & $\textbf{1.2020}\std{0.0392}$ \\
\bottomrule
\end{tabular}
\end{adjustbox}
\end{table*}

\begin{table*}[htp]
\caption{Performance(AUC) Comparison of Graph Transformer Models with RR Enhancement on dataset MolHIV}
\label{tab:transformer_rr_comparison}
\centering
\begin{adjustbox}{width=\textwidth}
\begin{tabular}{lccccc}
\toprule
\textbf{Model} & \textbf{Method} & \textbf{Cora} & \textbf{CiteSeer} & \textbf{PubMed} & \textbf{OGB-Arxiv} \\
\midrule
\multirow{2}{*}{Graphormer} & Original & 0.763 $\pm$ 0.012 & 0.691 $\pm$ 0.015 & 0.792 $\pm$ 0.008 & 0.718 $\pm$ 0.005 \\
 & + RR & \textbf{0.781} $\pm$ 0.011 & \textbf{0.705} $\pm$ 0.013 & \textbf{0.803} $\pm$ 0.007 & \textbf{0.729} $\pm$ 0.004 \\
%\addlinespace
\multirow{2}{*}{Graphormer with Spatial Encoding} & Original & 0.772 $\pm$ 0.010 & 0.702 $\pm$ 0.014 & 0.801 $\pm$ 0.007 & 0.725 $\pm$ 0.005 \\
 & + RR & \textbf{0.789} $\pm$ 0.009 & \textbf{0.715} $\pm$ 0.012 & \textbf{0.812} $\pm$ 0.006 & \textbf{0.736} $\pm$ 0.004 \\
%\addlinespace
\multirow{2}{*}{Graphormer with Graph Structure} & Original & 0.781 $\pm$ 0.011 & 0.712 $\pm$ 0.013 & 0.808 $\pm$ 0.007 & 0.732 $\pm$ 0.004 \\
 & + RR & \textbf{0.796} $\pm$ 0.010 & \textbf{0.724} $\pm$ 0.011 & \textbf{0.819} $\pm$ 0.006 & \textbf{0.742} $\pm$ 0.003 \\
\bottomrule
\end{tabular}
\end{adjustbox}
\end{table*}
\begin{table*}[htp]
\caption{Performance of Graph Transformer Networks (GT) with RR Enhancement}
\label{tab:GT_rr_comparison}
\centering
\begin{adjustbox}{width=0.8\textwidth}
\begin{tabular}{lcccc}
\toprule
\textbf{Node Classification} &{F1 Score } &{ACM} &{DBLP}& {IMDB}  \\
\midrule
\multirow{2}{*}{Base GT} 
& Original & 0.912 $\pm$ 0.014 & 0.938 $\pm$ 0.034 & 0.609 $\pm$ 0.023\\
& + RR & \textbf{0.923} $\pm$ 0.012 & \textbf{0.942} $\pm$ 0.025 & \textbf{0.724} $\pm$ 0.036\\
%\addlinespace
\bottomrule
\end{tabular}
\end{adjustbox}
\end{table*}

\begin{table*}[ht]

\centering
\begin{adjustbox}{width=\textwidth}
\begin{tabular}{|l|cc|ccc|ccc|ccc|}
\hline
\multirow{2}{*}{GNN Model on Watts-Strogatz} & \multirow{2}{*}{Data Load (s)} & \multirow{2}{*}{Load GPU (MB)} & \multicolumn{3}{c|}{GraphConv} & \multicolumn{3}{c|}{SAGEConv} & \multicolumn{3}{c|}{GATS} \\
%\cmidrule{4-12} \hline
  &  &  & train & val & GPU space & train & val & GPU space & train & val & GPU space \\ \hline
% -- Graph size 100 --
GAE (100)     & 0.30  & 0.73  & 0.85s & 0.31s & 0.21MB   & 0.70s & 0.23s & 0.38MB   & 0.97s & 0.38s & 0.22MB \\ 
DiGAE (100)   & 0.30  & 0.73  & 1.08s & 0.45s & 0.41MB   & 1.24s & 0.40s & 0.74MB   & 1.16s & 0.45s & 0.43MB \\ \hline
% -- Graph size 1000 --
GAE (1000)    & 1.40  & 7.31  & 1.91s & 0.74s & 0.58MB   & 4.71s & 0.27s & 0.75MB   & 2.05s & 0.72s & 0.59MB \\ 
DiGAE (1000)  & 1.40  & 7.31  & 3.06s & 0.56s & 1.11MB   & 2.24s & 0.77s & 1.44MB   & 2.05s & 0.60s & 1.13MB \\ \hline
% -- Graph size 10000 --
GAE (10000)   & 18.62 & 73.13 & 2.21s & 0.83s & 4.29MB   & 1.88s & 0.72s & 4.45MB   & 3.15s & 1.02s & 4.30MB \\ 
DiGAE (10000) & 18.62 & 73.13 & 2.90s & 1.06s & 8.12MB   & 3.45s & 0.97s & 8.45MB   & 2.75s & 1.13s & 8.13MB \\ \hline
% -- Graph size 50000 --
GAE (50000)   & 98.39 & 366.07 & 2.74s & 0.75s & 20.93MB  & 2.19s & 0.58s & 20.77MB  & 2.75s & 0.89s & 20.77MB \\ 
DiGAE (50000) & 98.39 & 366.07 & 3.34s & 0.91s & 39.57MB  & 3.38s & 1.27s & 39.25MB  & 5.25s & 1.74s & 39.26MB \\ \hline
% -- Graph size 100000 --
GAE (100000)  & 196.39& 731.48& 2.48s & 0.91s & 42.25MB  & 2.00s & 0.80s & 41.82MB  & 4.37s & 1.12s & 42.76MB \\ 
DiGAE (100000)& 196.39& 731.48& 3.90s & 1.18s & 78.60MB  & 2.93s & 1.05s & 78.51MB  & 6.22s & 2.24s & 80.07MB \\
\hline
\end{tabular}
\end{adjustbox}
\caption{Performance comparison of different GNN models on Watts-Strogatz graphs including data loading overhead.}
\label{t-ews}
\end{table*}

\subsection{Background of Graph Autoencoder}
\label{background}

%\begin{equation}\label{eq: interval RRC}
%\begin{split}
%   C_{ab} =  \Big[ &\hat{W}^{\alpha/2}_{ab} - d^{\textrm{RR}} \big|\hat{W}^{1-\alpha/2}_{ab} - \hat{W}^{\alpha/2}_{ab} \big|,  \\
%   & \hat{W}^{1-\alpha/2}_{ab} + d^{\textrm{RR}}\big|\hat{W}^{1-\alpha/2}_{ab} - \hat{W}^{\alpha/2}_{ab}\big| \Big], \; 
% \end{split}
% \end{equation}
 
%\begin{equation}
%   (a, b) \in \displaystyle \mE^{test},
%\end{equation}

%for conformal prediction-based RR (CP-RR), and for conformal quantile regression-based RR (CQR-RR),

\subsubsection{Guaranteed Edge Weight Prediction Using GNNs}\label{sec: problem}
Let $\gG=(\sV,\mE)$ be a graph with node set $V$ and edge set $E \subseteq \sV \times \sV$.
Assume the graph has $n$ nodes with $f$ features.
Let $\mX \in \displaystyle R^{n\times m}$ be the node feature matrix, and $X_{i,:} \in \displaystyle R^{f}$ be the feature vector of the $i^{th}$ node. 
The binary adjacency matrix of $\gG$, $ \mA \in \{0, 1\}^{n\times n}$, encodes the binary (unweighted) connecting structure of the graph. It is defined by:
\begin{equation} 
A_{ij} =
\begin{cases}
1, & \textrm{if } (i, j) \in \mE; \\
0, & \textrm{otherwise}.
\end{cases}
\end{equation}
Then, we define the weight matrix as $\mW \in \displaystyle R^{n\times n}{\geq 0}$, where $W_{ij}$ denotes the weight rather than the binary of the edge connecting node $i$ to node $j$. In the context of a road system, for example, we can interpret $W_{ij}$ as the volume of traffic transitioning from junction $i$ to junction $j$.

We partition the edge set $\mE$ into three disjoint subsets: $\mE^{train}$, $\mE^{val}$,$\mE^{calib}$ and $\mE^{test}$, while satisfy that $\mE = \mE^{train} \cup \mE^{val} \cup \mE^{calib} \cup \mE^{test}$.
We assume that the weights of the edges in $\mE^{train}$ and $\mE^{val}$ are known.
The objective is to estimate the unknown weights of the edges in $\mE^{test}$.
Additionally, we assume that the entire graph structure, represented by the adjacency matrix $A$, is known.

To mask the validation and test sets, we define \begin{equation}
\mA^{train} \in \{0, 1\}^{n\times n}, \quad 
A^{train}_{ij} =
\begin{cases}
1, & \textrm{if } (i, j) \in \mE^{train}; \\
0, & \textrm{otherwise}.
\end{cases}
\end{equation}
$\mA^{val}$, $\mA^{calib}$ and $\mA^{test}$ are defined  in the same way based on $\mE^{val}$, $\mE^{calib}$ and $\mE^{test}$, respectively.

If $(i, j) \notin \mE^{train}$,  it is possible to assign a small positive number to the corresponding element $Wtrain_{ij}$, such as an assigned value or the minimum of the existing edge weights. This can represent prior knowledge or assumptions about the unknown edge weight. In the following part, we use a positive constant $\delta > 0$ to represent this minimal or assigned value. Incorporating this unknown edge weight information effectively leverages the underlying graph structure. The resulting weighted adjacency matrix is:
\begin{equation}\label{eq: weighted adj train3}
\mW^{train} =
\begin{cases}
W_{ij}, & \textrm{if } (i, j) \in \mE^{train}; \\
\delta, & \textrm{if } (i, j) \in \mE^{val} \cup \mE^{calib} \sup \mE^{test}; \\
0, & \textrm{otherwise},
\end{cases}
\end{equation}

In the transductive setting (details can been seen in appendix Figure~\ref{fig: transductive}(a)), the structure of the entire graph, represented by the adjacency matrix $\mA$, is known during the training, validation, and testing phases. To calibrate the prediction, we extract a subset from $\mE^{test}$ as a calibration edge set. This ensures that the calibration and test samples are exchangeable.

Consider edge weight prediction in traffic networks. The road system, $\mA$, and partial traffic volumes, $\mW^{train} + \mW^{val}$, are known. The task is to predict volumes, $\mW^{test}$, for the remaining roads.

During training, models observe node features and graph structure to learn functions for node classification/regression and embedding. At inference, models deduce edge connections between nodes (Figure~\ref{fig: transductive}).

Three GNN approaches are evaluated. The first is a Graph Autoencoder (GAE) \cite{kipf2016variational} that trains and infers on the full graph. The second is DiGAE \cite{kollias2022directed}, a directed GAE variant. The third is the line graph neural network (LGNN) \cite{cai2021line} that transforms edges to nodes in line graphs.
\subsection{Appendices Figures}
\subsubsection{Schematic figure for transductive and inductive settings for link prediction.}
\begin{figure*}[ht]
\centerline{\includegraphics[height=3.5cm, width=0.7\textwidth,clip=]{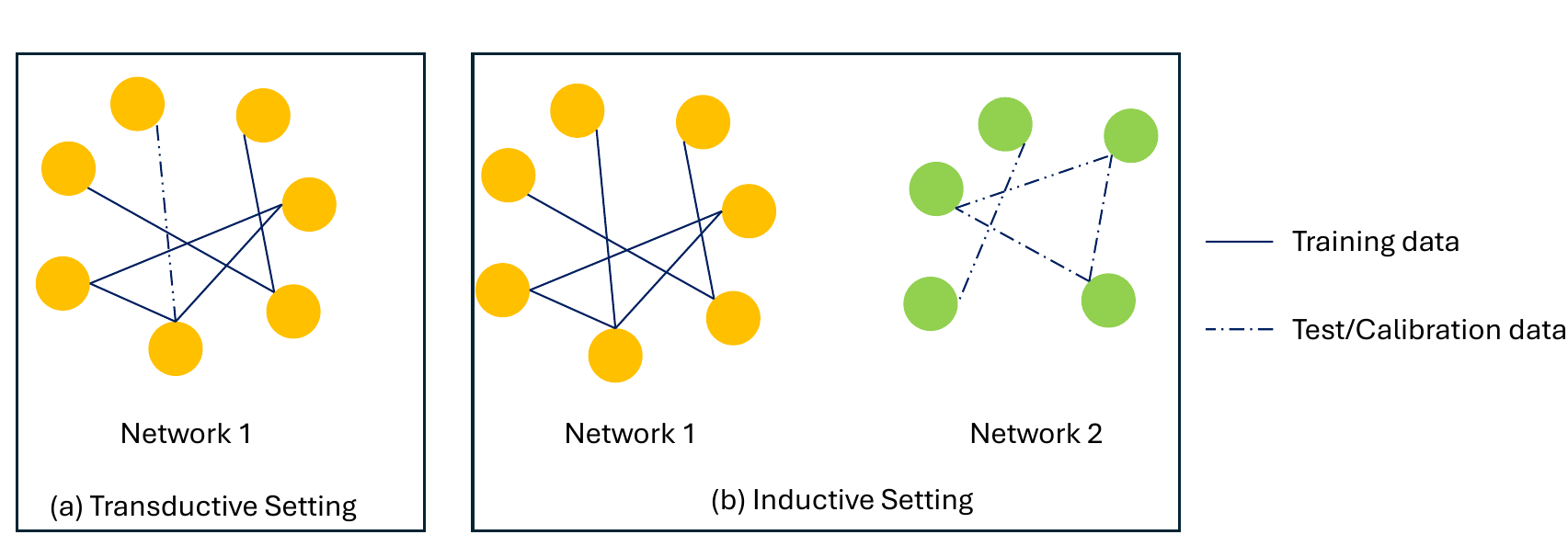}}
\caption{Schematic figure for transductive and inductive settings for edge weight prediction.
Different colors indicate the availability of the nodes during the training or testing phases.
Solid and dashed lines represent edges used for training and the predicted edge in the testing phases, respectively.
(a) Transductive edge weight prediction performs both training and inference on the same graph.
(b) Inductive edge weight prediction inference is performed on a new, unseen graph.}
\label{fig: transductive}
\end{figure*}

There are two major braches in link prediction problem settting: transductive and inductive settings. Schematic figure for transductive and inductive settings for edge weight prediction are shown in Figure~\ref{fig: transductive}
\subsubsection{Learning the error structure}
\begin{figure}[ht]
\centering
\includegraphics[height=4.2cm]{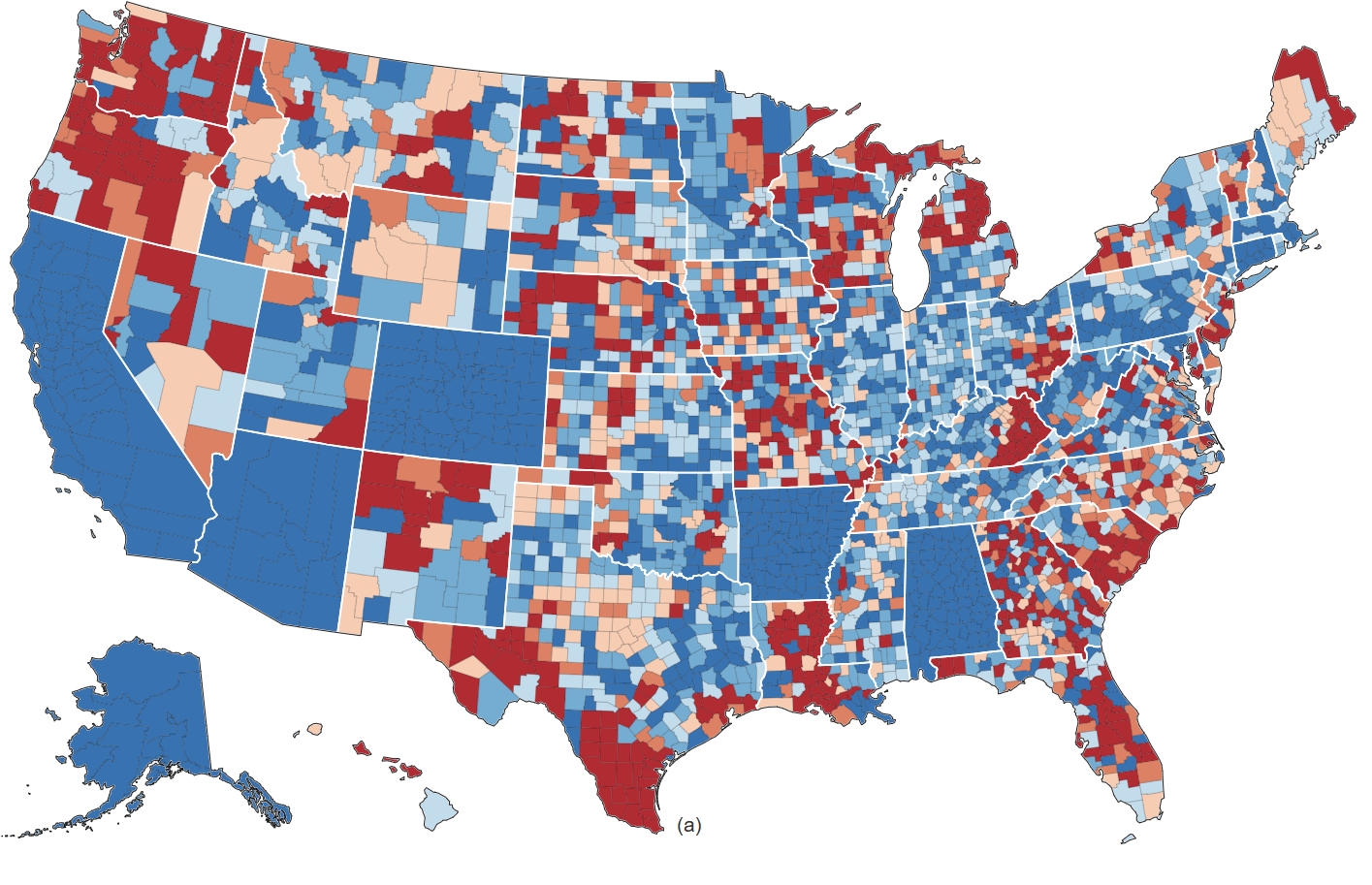}
\includegraphics[height=4.2cm]{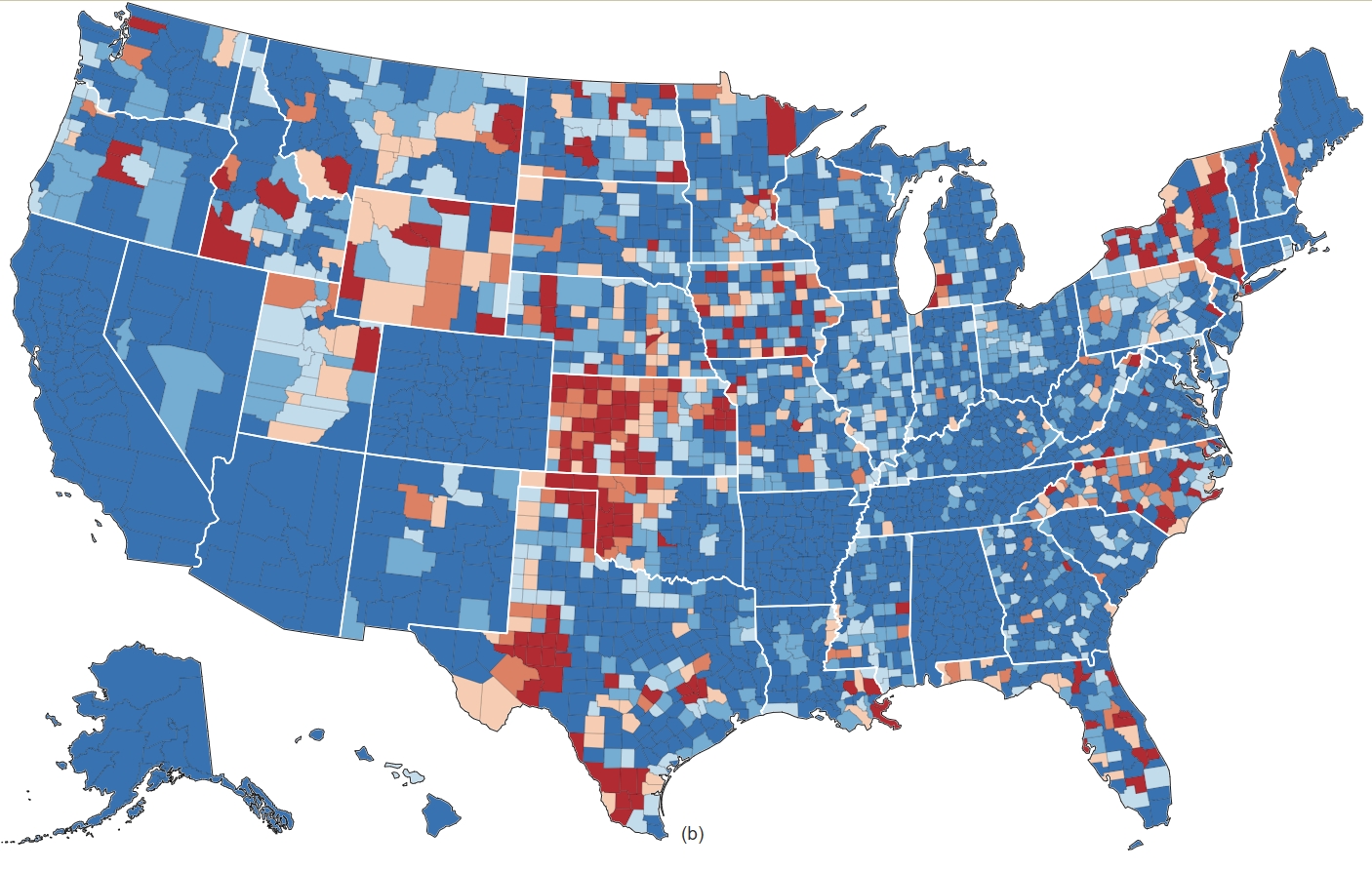}
\caption{Residual for predicted 2016 U.S. county-level presidential election. (a) The baseline model's residual \cite{jia2020residual}, which is the normalized absolute difference between the predicted and ground truth vote count.  (b) RR-GNN's residual result. The results indicate that the proposed RR-GNN achieves smaller and more uniform error/residuals.}
\label{fig: comparison2}
\end{figure}

\begin{figure*}[ht]
\centering
\includegraphics[height=3.4cm,width=0.244\textwidth]{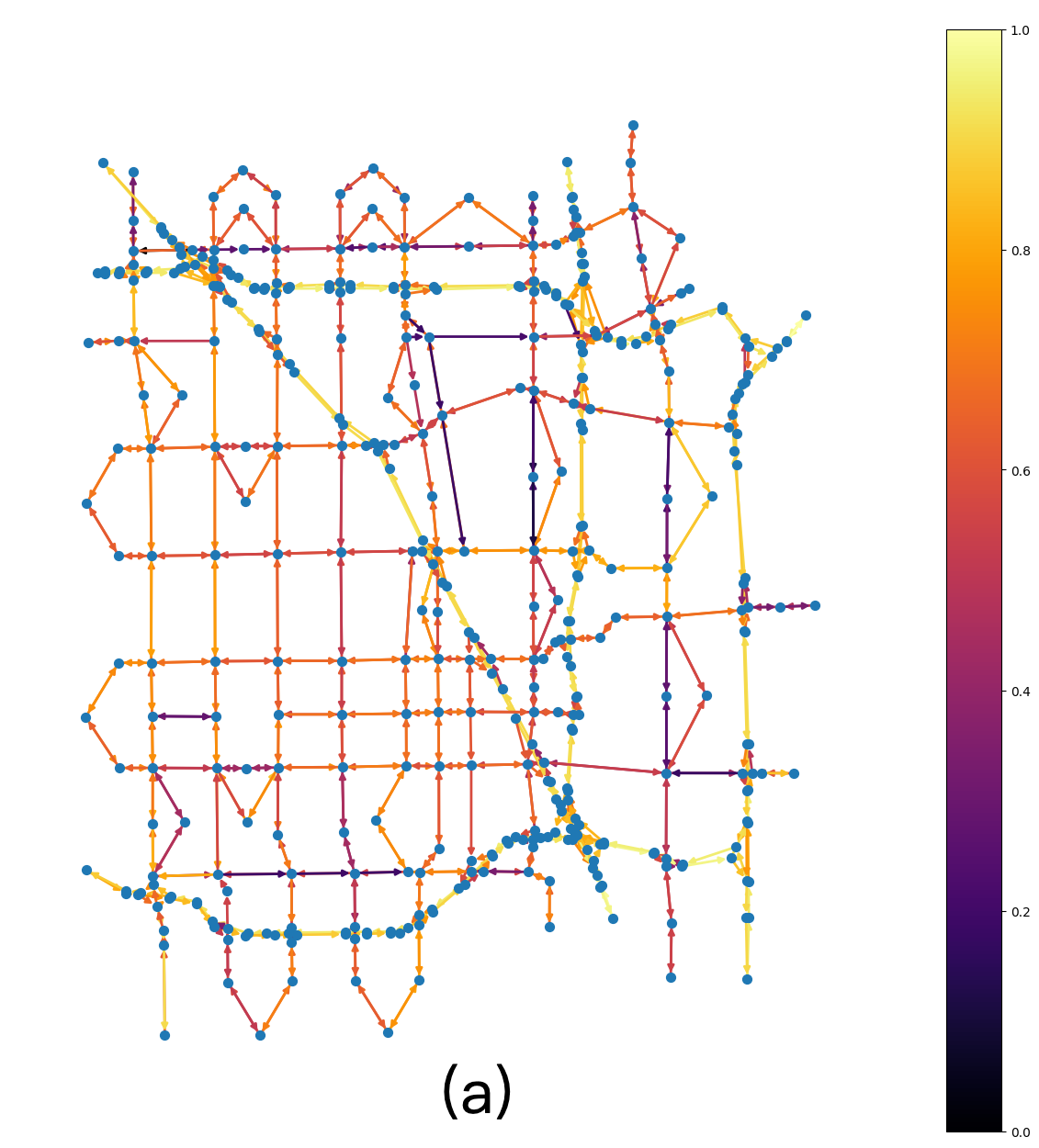}
\includegraphics[height=3.4cm,width=0.244\textwidth]{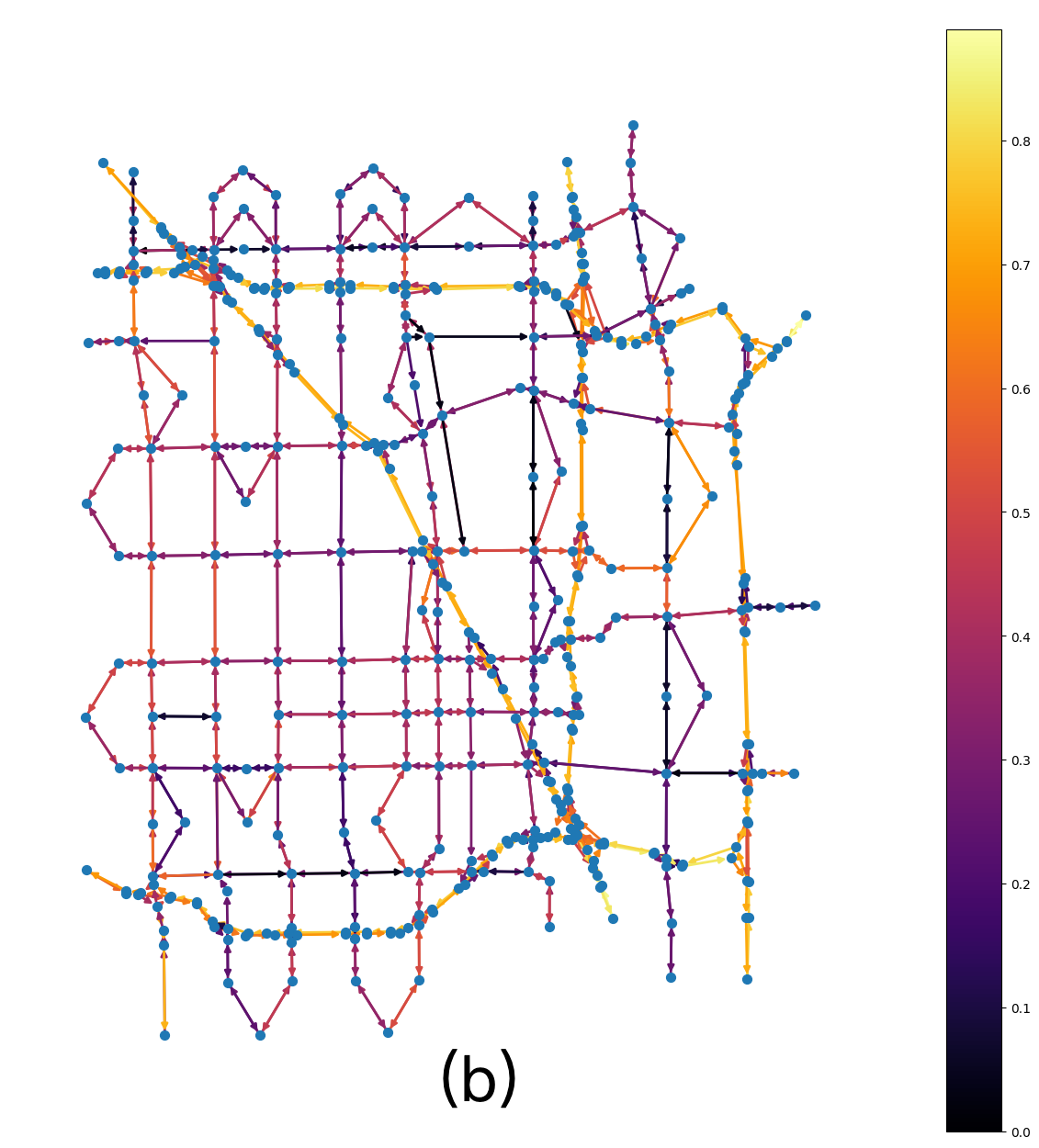}
\includegraphics[height=3.4cm,width=0.244\textwidth]{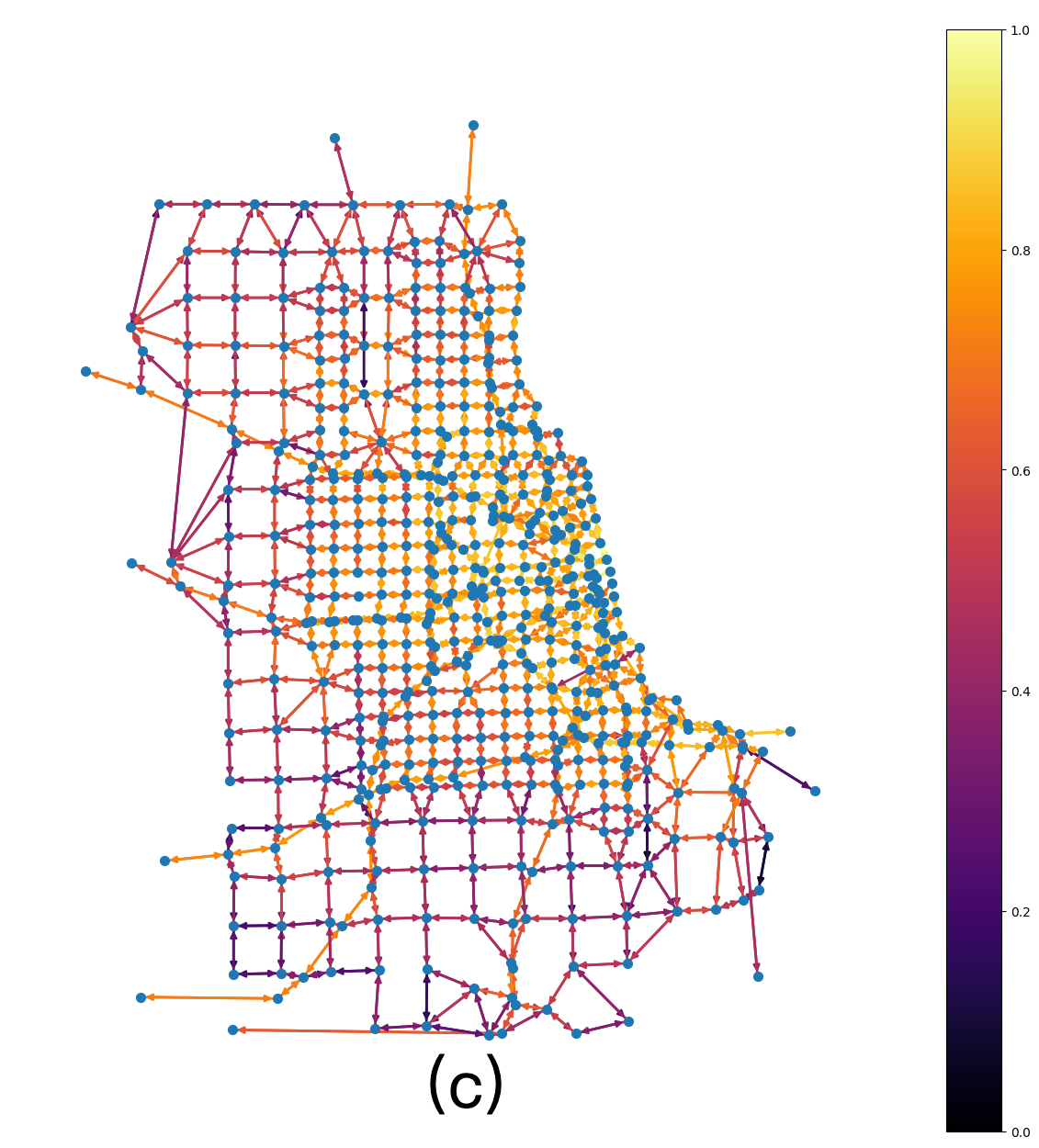}
\includegraphics[height=3.4cm,width=0.244\textwidth]{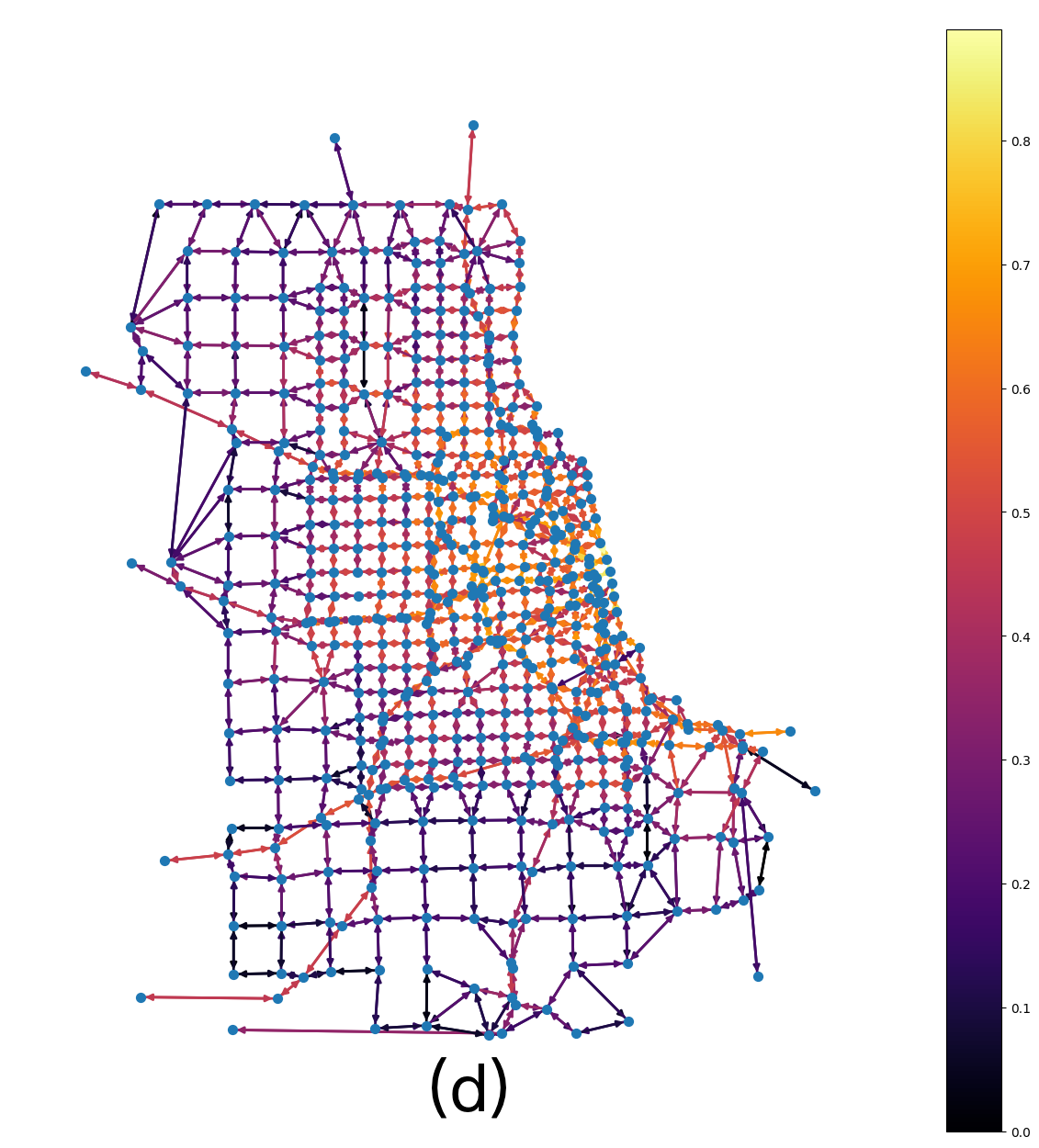}
\caption{Residual between predicted and actual traffic volumes across roads in two cities under different models. We took the average volume at the start and end points as the road traffic. (a) Residual of predicted roads' traffic volume in Anaheim of baseline model \cite{huang2024uncertainty}, which is the absolute value of prediction and ground truth. (b) Residual of predicted roads' traffic volume in Anaheim of RR-GAE. (c) Residual of predicted roads' traffic volume in Chicago of baseline model. (d) Residual of predicted roads' traffic volume in Chicago of RR-GAE. For each city, the residuals from the two models were independently normalized to a 0-1 range for comparison purposes.}
\label{fig:comparisontraffic}
\end{figure*}

In order to show our method's outperformance in learning the error structure in real-world applications, we tested our method in two datasets: 2016 U.S. county-level presidential election and Traffic volums prediction in Anaheim and Chicago transportation networks. Results are shown in Figure ~\ref{fig: comparison2} and Figure~\ref{fig:comparisontraffic}.

\begin{algorithm}
\caption{Residual Reweighted Conformalized Graph Neural Network for Node Regression}
\label{alg: nodd regression}
\textbf{Input:} The binary adjacency matrix $\mA \in \{0, 1\}^{n\times n}$, training node features $\mX\in \displaystyle R^{n\times m}$, training node set $\sV^{train}$ and label , $\displaystyle \vy^{train}$, validation node set $\sV^{val}$ and label- $\displaystyle \vy^{val}$ (Used for training Residual GNN), calibration nodes $\sV^{calib}$ and label $\displaystyle \vy^{calib}$, and test nodes $\sV^{test}$, user-specified error rate $\alpha \in (0,1)$, two GNN model $\gG_{\vtheta_1}$ and $\gG_{\vtheta_2}$ with trainable parameter $\displaystyle \vtheta_1$ and $\displaystyle \vtheta_2$.\\
\begin{algorithmic}[Ht]
\STATE Train the model $g_{\vtheta_1}$ and $g_{\vtheta_2}$ with $\displaystyle \vy^{train}$ and $\displaystyle \vy^{val}$ according to Algorithm 2 in main paper. %\ref{closs_train}.
\STATE Compute the nonconformity score, which quantifies the interval the predicted calibration node labels:
\begin{equation}
V^{\textrm{RR}}_{i} = \max\left \{ \frac{\hat{y}^{\alpha/2}_{i} - y^{calib}_{i}}{\big|\hat{R}_{i}\big|}, \frac{y^{calib}_{i} - \hat{y}^{1-\alpha/2}_{i}}{\big|\hat{R}_{i}\big|}  \right\}, \; 
\label{eq-nr}
\end{equation}
\begin{equation}
i \in {\displaystyle \sV}^{calib},    
\end{equation}
\STATE Compute $d =$ the $k$th smallest value in $\{V^{\textrm{RR}}_{i}\}$, where $k=\lceil(|{\displaystyle \sV}^{calib}| +1)(1-\alpha)\rceil$;
\STATE Construct a prediction interval for test nodes:  
\begin{equation}
    C_{a} = \Big[\hat{y}^{\alpha/2}_{a} - d\big|\hat{R}_{a}\big|, \hat{y}^{1-\alpha/2}_{a} + d\big|\hat{R}_{a}\big| \Big], \; a \in {\displaystyle \sV}^{test}. \nonumber
\end{equation}
\end{algorithmic}
\textbf{Output:} Prediction of confidence intervals for the test nodes $a \in {\displaystyle \sV}^{test}$ with the coverage guarantee:
\begin{equation}
    p\big(y^{test}_{a} \in C_{a} \big) \geq 1 - \alpha.
\end{equation}
\end{algorithm}

\subsection{Nonconformity Score and Evaluation Metrics}
{\bf Nonconformity Score}
For the nonconformity score in main paper is for conformal prediction-based RR (CP-RR).
On the other hand, for conformal quantile regression-based RR (CQR-RR),
The nonconformity score is

\begin{equation}\label{eq: interval RRC}
\begin{split}
   C_{ab} =  \Big[ &\hat{W}^{\alpha/2}_{ab} - d^{\textrm{RR}} \big|\hat{W}^{1-\alpha/2}_{ab} - \hat{W}^{\alpha/2}_{ab} \big|,  \\
   & \hat{W}^{1-\alpha/2}_{ab} + d^{\textrm{RR}}\big|\hat{W}^{1-\alpha/2}_{ab} - \hat{W}^{\alpha/2}_{ab}\big| \Big], \; 
 \end{split}
\end{equation}
\begin{equation}
   (a, b) \in \displaystyle \mE^{test},
\end{equation}

{\bf Evaluation Metrics:}
For evaluation, we use the marginal coverage, defined as 
\begin{equation}\label{eq: cover}
    \textrm{cover} = \frac{1}{|\displaystyle E^{test}|} \sum_{(i,j)\in \displaystyle 
    E^{test}} \mathbbm{1}\big(Wtest_{ij} \in C_{ij}\big),
\end{equation}
where $C_{ij}$ is prediction interval for edge $(i, j)$. Another one is inefficiency which is defined as
\begin{equation}\label{eq: ineff}
    \textrm{ineff} = \frac{1}{|\displaystyle E^{test}|} \sum_{(i,j)\in \displaystyle E^{test}} |C_{ij}|,
\end{equation}
which measures the average length of the prediction interval.

In addition to the marginal coverage, we also consider the conditional coverage. Specifically, we measure the coverage over a slab of the feature space $S_{\displaystyle v, a, b}= \big\{[\mX_{i, :} \mathbin\Vert \mX_{j, :}]\in \displaystyle R^{2f}: a \leq \displaystyle v^\top \mX \leq b \big\}$ \cite{romano2020classification, cauchois2020knowing}, where $[\mX_{i, :} \mathbin\Vert \mX_{j, :}]$ denotes the node feature of two connected nodes of an edge $(i, j)$ and $\displaystyle v \in \displaystyle R^{2f}$ and $a < b \in \displaystyle R$ are chosen adversarially and independently from the data. For any prediction interval $f_\theta^*$ and $\delta \in (0, 1)$, the \textit{worst slice coverage} is defined as 
\begin{equation}\label{eq: cond cover}
\begin{split}
    \textrm{WSC}(f_\theta^*, \delta) &= \inf\limits_{\substack{\displaystyle v \in \displaystyle R^{2f}, \\ a < b \in \displaystyle R}} \big \{ P \big( Wtest_{ij} \in C_{ij} \mid [\mX_{i, :} \mathbin\Vert \\ & \phantom{-----}  \mX_{j, :}] \in S_{\displaystyle v, a, b}  \big) \\
    & \phantom{---} \textrm{s.t. } P([\mX_{i, :} \mathbin\Vert \mX_{j, :}] \in S_{\displaystyle v, a, b}) \geq \delta  \big \}.
\end{split}
\end{equation}

\subsection{Algorithm for Node Classification}

\begin{algorithm}[ht]
\caption{Residual Reweighted Conformalized Graph Neural Network for Node Classification}
\label{alg: nodd classification}
 \textbf{Input:} The binary adjacency matrix $\mA \in \{0, 1\}^{n\times n}$, training node features $\mX\in \displaystyle \textbf{R}^{n\times m}$, training node and class label variable ${\sV}^{train}$, $\displaystyle \vl^{train}$ ($\vl^{train}$ is the one-hot of $\vl^{train}$. The situation is also met for $\hat{l}$ and $\hat{l}$ which are class output of model), validation nodes and label ${\sV}^{val}$,  $\displaystyle \vl^{val}$ (Used for training Residual GNN), calibration nodes and label ${\sV}^{calib}$, $\vl^{calib}$, and test nodes ${\sV}^{test}$, user-specified error rate $\alpha \in (0,1)$, two GNN model $g_{\vtheta_1}$ and $g_{\vtheta_2}$ with trainable parameter $\vtheta_1$ and $\vtheta_2$.\\
\begin{algorithmic}[Ht]
\STATE Train the model $g_{\vtheta_1}$ and $g_{\vtheta_2}$ with $\displaystyle \vl^{train}$ and $\displaystyle \vl^{val}$ according to Algorithm 2 in main paper %\ref{closs_train}.
\STATE Compute the score which quantifies the residual of the calibration node classes $\displaystyle \vl^{calib}$ projected onto the nearest quantile produced by $g_{\vtheta_1}$ and $g_{\vtheta_2}$ :
\begin{equation}
    V^{\textrm{RR}}_{i} = \max\left \{ \frac{\big|\hat{l}^{\alpha/2}_{i} - l^{calib}_{i}\big|}{\big|\hat{\mathbf{R}}_{i}\big| + \epsilon}, 
    \frac{\big|l^{calib}_{i} - \hat{l}^{1-\alpha/2}_{i}\big|}{\big|\hat{\mathbf{R}}_{i}\big| + \epsilon}  \right\}, \;
    \label{eq-nc}
\end{equation}
\begin{equation}
    i \in \sV^{calib},
\end{equation}
\STATE Compute $d =$ the $k$th smallest value in $\{V^{\textrm{RR}}_{i}\}$, where \
\begin{equation} \label{diffquantile}
    k=DiffQuantile(\lceil(| \sV^{calib} | +1)(1-\alpha)\rceil);
\end{equation}
\STATE Construct a prediction interval for test nodes:  
\begin{equation}
    C_{a} = \Big[\hat{l}^{\alpha/2}_{a} - d_{(m)}\big|\hat{R}_{a}\big|, \hat{l}^{1-\alpha/2}_{a} + d_{(m)}\big|\hat{R}_{a}\big| \Big], \; a \in \sV^{test}. \nonumber
\end{equation}
\end{algorithmic}
\textbf{Output:} Prediction of confidence intervals for the test nodes $(a, b) \in \sV^{test}$ with the coverage guarantee:
\begin{equation}
    p\big(l^{test}_{a} \in C_{a} \big) \geq 1 - \alpha.
\end{equation}
\end{algorithm}
\subsection{Complete table of experiment}

\subsection{Theoretical guarantee of our method}
7.1. Conformal Prediction

We assume we have access to the graph structure, $A$, the node features, $X$, and the weighted adjacency matrix $W^{\text {train }}(3)$. Let $(a, b)$ be the endpoints of a test edge. We aim to generate a prediction interval, $C_{a b}=\left(f_\theta\left((a, b), A, X, W^{\text {train }}\right)\right) \subset$ $\mathbb{R}$, for the weight of the such a test edge. The prediction interval should be marginally valid, i.e. it should obey
\begin{equation}
P\left(W_{a b} \in C_{a b}\right) \geq 1-\alpha,
\label{equation-cp}
\end{equation}
where $\alpha \in(0,1)$ is a user-defined error rate. The probability is over the datagenerating distribution. For efficiency, we focus on the split CP approach\cite{papadopoulos2002inductive}, using the training edge set $E^{\text {train}}$ for training and the calibration edge set $E^{\text {calib }}$ for calibration. $E^{\text {train }}$ is used to fit the prediction model, $f_\theta$, and a conformity score is calculated for each sample in $E^{\text {calib }}$. The conformity score evaluates how well the predictions match the observed labels.

Proposition 1. The prediction intervals generated by split CP (Algorithm 1), CQR (Algorithm 2), and RR are marginally valid, i.e. obey equation~\ref{equation-cp} in appendices material.

Proof. First, we show that the calibration and test conformity scores defined in equation~\ref{eq-nr} and ~\ref{eq-nc} in appendices material and (12) in main paper are exchangeable. Given the entire graph structure, $A$, all the node features, $X$, and the edge weights of the training edges, $W^{\text {train }}$, the node embeddings are trained based on $W^{\text {train }}$, and the edge weights in the remaining $E^{\text {ct }}$ are set randomly, the division of $E^{\text {ct }}$ into $E^{\text {calib }}$ and $E^{\text {test }}$ have no impact on the training process. Consequently, the conformity scores for $E^{\text {calib }}$ and $E^{\text {test }}$ are exchangeable. In practice, we split $E^{\text {ct }}$ into $E^{\text {calib }}$ and $E^{\text {test }}$ randomly (as detailed in Section ~\ref{background} in appendices material) by converting the graph into its line graph and then selecting nodes uniformly at random.

We also explore an alternative proof which is equivalent to the proof in Huang et al.\cite{huang2024uncertainty} but applied within a line graph setting. Consider the original graph $G=(V, E)$ and its corresponding line graph $G^{\prime}=\left(V^{\prime}, E^{\prime}\right)$, where $V^{\prime}=E$ and $E^{\prime}$ denotes adjacency between edges in G. After randomly dividing E into $\mathbb{E}^{\text {train}}$ and $E^{\mathrm{ct}}$, and further splitting $E^{\mathrm{ct}}$ into $E^{\text {calib }}$ and $E^{\text {test }}$, the edges of $G$ transforms into nodes in $G^{\prime}$. This setup mirrors the node division in the line graph. We train node embeddings on $E^{\text {train }}$ using a graph autoencoder, which aligns with fixing the training node set in $G^{\prime}$. Given this fixed training set, any permutation and division of $E^{\text {ct }}$ (which corresponds to nodes in $G^{\prime}$ ) doesn't affect the training, and thus the conformity scores computed for $E^{\text {calib }}$ and $E^{\text {test }}$ are exchangeable.

Given this exchangeability of nonconformity scores, the validity of the prediction interval produced by CP and CQR follows from Theorem 2.2 of Lei et al.\cite{lei2018distribution} and Theorem 1 of Romano, Patterson, and Candes\cite{romano2019conformalized}. Let $V$ be the conformity score of CQR. The RR approach performs a monotone transformation of $V$, defined as
$$
\Phi_{i j}(V)=\frac{V}{\left|\hat{W}_{i j}^{1-\alpha / 2}-\hat{W}_{i j}^{\alpha / 2}\right|},
$$

where $i$ and $j$ are two nodes in the $\operatorname{graph}^2$. For all $(i, j)$ and all $V, \Phi_{i j}^{\prime}(V)=$ $\frac{\partial \Phi_{i j}(V)}{\partial V}>0$, i.e. the transformation is strictly monotonic in $V$. This implies $\Phi_{a b}$ is invertible for any test edge, $(a, b)$. Let $\Phi_{a b}^{-1}$ be the inverse of $\Phi_{a b}$. The Inverse Function Theorem implies $\Phi_{a b}^{-1}$ is also strictly increasing. Now suppose $d^{\mathrm{RR}}$ is the $k$ th smallest value in $\left\{V_{i j}^{\mathrm{RR}}\right\}=\left\{\Phi_{i j}\left(V_{i j}\right)\right\}, k=\left\lceil\left(\left|E^{\text {calib }}\right|+1\right)(1-\alpha)\right\rceil$. Then for a test edge $(a, b)$,
$$
P\left(\Phi_{a b}\left(V_{a b}\right) \leq d^{\mathrm{RR}}\right)=\frac{\left\lceil\left(\left|E^{\text {calib }}\right|+1\right)(1-\alpha)\right\rceil}{\left|E^{\text {calib }}\right|+1} \geq 1-\alpha .
$$

Using the monotonicity of $\Phi_{a b}^{-1}$,
$$
\begin{aligned}
1-\alpha & \leq P\left(\Phi_{a b}\left(V_{a b}\right) \leq V_k^{\mathrm{RR}}\right) \\
& =P\left(V_{a b} \leq \Phi_{a b}^{-1}\left(d^{\mathrm{RR}}\right)\right) \\
& =P\left(W_{a b} \in C_{a b}\right)
\end{aligned}
$$

The final equation is derived from the construction of the prediction interval in $C_a$ in Algorithm 1 in main paper and Algorithm 2 and 3 in appendices material and the validity of CQR. This shows that the prediction intervals based on the reweighted conformity scores are valid.              Proof done.

\begin{table*}[ht]
\caption{Results of RR-GNN on Node Regression Datasets}
\label{tab:eff_all_models12-2}
\centering
\begin{adjustbox}{width=\textwidth}
\begin{tabular}{|l|c|c|c|c|c|c|c|c|}
\toprule
Dataset& \multicolumn{2}{c|}{GraphSAGE} & \multicolumn{2}{c|}{SGC}  & \multicolumn{2}{c|}{GCN} & \multicolumn{2}{c|}{GATS} \\ \cmidrule{1-9}
 Metrics & cover$^x$  & ineff & cover$^x$ & ineff & cover$^x$ & ineff& cover$^x$ & ineff\\\midrule
Anaheim: CF-GNN&$0.9520\std{0.0669} $ 
&$\textbf{1.9231}\std{0.0483}$
&$0.9559\std{0.0617}$
&$2.2031\std{0.0241}$
&$0.9519\std{0.0531} $ 
&$2.3782\std{0.0533}$
&$0.9523\std{0.0302}$
&$2.1499\std{0.0463}$\\
Anaheim: Cluster-GNN&$0.9532\std{0.042}$
&$1.8954\std{0.037}$
&$0.9561\std{0.035}$
&$2.1423\std{0.031}$
&$0.9528\std{0.041}$
&$2.2451\std{0.029}$
&$0.9541\std{0.028}$
&$2.0321\std{0.025}$\\
Anaheim: RR-GAE&$0.9539\std{0.038}$
&$1.8732\std{0.032}$
&$\textbf{0.9567}\std{0.031}$
&$2.0987\std{0.028}$
&$0.9532\std{0.036}$
&$2.1934\std{0.026}$
&$0.9563\std{0.024}$
&$1.9623\std{0.022}$\\
Anaheim: Clsuter-RR-GAE&$\textbf{0.9543}\std{0.0320} $ 
&${1.9647}\std{0.0197}$
&$0.9577\std{0.0657}$
&$\textbf{2.0188}\std{0.0246}$
&$\textbf{0.9585}\std{0.0413}$ 
&$\textbf{2.2179}\std{0.0254}$
&$\textbf{0.9638}\std{0.0302}$
&$\underline{\textbf{1.8996}}\std{0.0249}$\\
\midrule
Chicago: CF-GNN&$0.9448\std{0.0519} $ 
&${2.3426}\std{0.0384}$
&$0.9486\std{0.0247}$
&$1.0423\std{0.0372}$
&$0.9505\std{0.0447}$
&$2.0456\std{0.0443}$
&$0.9508\std{0.0569}$
&$1.1396\std{0.0686}$\\
Chicago: Cluster-GNN&$0.9461\std{0.039}$
&$2.2894\std{0.034}$
&$0.9492\std{0.031}$
&$1.1895\std{0.029}$
&$0.9513\std{0.037}$
&$1.8742\std{0.031}$
&$0.9516\std{0.042}$
&$1.1254\std{0.045}$\\
Chicago: RR-GAE&$0.9472\std{0.035}$
&$2.2673\std{0.029}$
&$\textbf{0.9498}\std{0.028}$
&$1.2567\std{0.026}$
&$0.9519\std{0.033}$
&$1.6923\std{0.027}$
&$0.9519\std{0.038}$
&$1.1489\std{0.039}$\\
Chicago: Cluster-RR-GAE&$\textbf{0.9476}\std{0.0426} $ 
&$\textbf{2.2291}\std{0.0325}$
&$0.9546\std{0.0328}$
&$\textbf{1.2012}\std{0.0250}$
&$\textbf{0.9538}\std{0.0356}$ 
&$\underline{\textbf{1.5769}}\std{0.0252}$
&$\textbf{0.9540}\std{0.0362}$  
&$1.1283\std{0.0256}$\\ \midrule  
Education: CF-GNN&$0.9501\std{0.0242} $ 
&${2.3808}\std{0.0427}$
&$0.9500\std{0.0285}$
&$2.4892\std{0.0351}$
&$0.9483\std{0.0408}$
&$2.4380\std{0.0452}$
&$0.9502\std{0.0392}$
&$2.4209\std{0.0376}$\\
Education: Cluster-GNN&$0.9513\std{0.031}$
&$2.3145\std{0.038}$
&$0.9517\std{0.033}$
&$2.3721\std{0.032}$
&$0.9496\std{0.035}$
&$2.2894\std{0.034}$
&$0.9518\std{0.036}$
&$2.3256\std{0.033}$\\
Education: RR-GAE&$0.9529\std{0.029}$
&$2.1932\std{0.027}$
&$0.9534\std{0.030}$
&$2.1478\std{0.028}$
&$0.9508\std{0.032}$
&$2.0321\std{0.029}$
&$0.9532\std{0.031}$
&$2.1423\std{0.030}$\\
Education: Cluster-RR-GAE&$\textbf{0.9599}\std{0.0417} $ 
&$\textbf{2.0573}\std{0.0280}$
&$\textbf{0.9586}\std{0.0225}$
&$\textbf{2.0445}\std{0.0239}$
&$\textbf{0.9580}\std{0.0333}$ 
&$\textbf{1.8731}\std{0.0260}$
&$\textbf{0.9594}\std{0.0386}$  
&$\underline{\textbf{1.9075}}\std{0.0221}$\\ \midrule
Election: CF-GNN&$0.9498\std{0.0211} $ 
&${0.9268}\std{0.0429}$
&$0.9495\std{0.0215}$
&$0.9279\std{0.0302}$
&$0.9506\std{0.0473}$
&$0.9009\std{0.0282}$
&$0.9488\std{0.0363}$
&$0.9136\std{0.0681}$\\
Election: Cluster-GNN&$0.9503\std{0.028}$
&$0.9152\std{0.038}$
&$0.9501\std{0.027}$
&$0.9124\std{0.035}$
&$0.9512\std{0.041}$
&$0.8723\std{0.031}$
&$0.9496\std{0.033}$
&$0.8945\std{0.042}$\\
Election: RR-GAE&$0.9509\std{0.025}$
&$0.9037\std{0.029}$
&$0.9523\std{0.024}$
&$0.8956\std{0.028}$
&$0.9518\std{0.036}$
&$0.8234\std{0.026}$
&$0.9514\std{0.030}$
&$0.8562\std{0.035}$\\
Election: Cluster-RR-GAE&$\textbf{0.9558}\std{0.0215}$ 
&$\textbf{0.9213}\std{0.0279}$
&$\textbf{0.9567}\std{0.0242}$
&$\textbf{0.9487}\std{0.0259}$
&$\textbf{0.9510}\std{0.0432}$ 
&$\textbf{0.9343}\std{0.0341}$
&$\textbf{0.9567}\std{0.0317}$  
&$\underline{\textbf{0.6698}}\std{0.0201}$\\ \midrule
Income: CF-GNN&$0.9512\std{0.0264} $ 
&${2.7580}\std{0.0342}$
&$0.9504\std{0.0405}$
&$2.4892\std{0.0302}$
&$0.9511\std{0.0250}$
&$2.5272\std{0.0318}$
&$0.9508\std{0.0329}$
&$2.4396\std{0.0328}$\\
Income: Cluster-GNN&$0.9521\std{0.035}$
&$2.6723\std{0.041}$
&$0.9513\std{0.038}$
&$2.3721\std{0.037}$
&$0.9526\std{0.033}$
&$2.4189\std{0.036}$
&$0.9519\std{0.034}$
&$2.3254\std{0.035}$\\
Income: RR-GAE&$0.9538\std{0.032}$
&$2.5342\std{0.038}$
&$\textbf{0.9524}\std{0.036}$
&$2.1423\std{0.034}$
&$0.9539\std{0.031}$
&$2.1932\std{0.033}$
&$0.9527\std{0.033}$
&$2.1567\std{0.032}$\\
Income: Cluster-RR-GAE&$\textbf{0.9552}\std{0.0618}$ 
&$\textbf{2.1003}\std{0.0492}$
&$0.9519\std{0.0513}$
&$\textbf{1.9616}\std{0.0358}$
&$\textbf{0.9566}\std{0.0501} $ 
&$\textbf{1.9203}\std{0.0354}$
&$\textbf{0.9545}\std{0.0347}$  
&$\underline{\textbf{1.8555}}\std{0.0423}$\\ \midrule
Unemploy: CF-GNN&$0.9526\std{0.0415}$ 
&${2.2298}\std{0.0523}$
&$0.9510\std{0.0320}$
&$2.4587\std{0.0491}$
&$0.9506\std{0.0294}$
&$2.5013\std{0.0326}$
&$0.9502\std{0.0354}$
&$2.4332\std{0.0376}$\\
Unemploy: Cluster-GNN&$0.9531\std{0.038}$
&$2.1932\std{0.045}$
&$0.9519\std{0.036}$
&$2.3256\std{0.042}$
&$0.9513\std{0.034}$
&$2.3721\std{0.038}$
&$0.9516\std{0.033}$
&$2.2894\std{0.039}$\\
Unemploy: RR-GAE&$0.9542\std{0.035}$
&$2.1423\std{0.039}$
&$0.9524\std{0.033}$
&$2.1932\std{0.036}$
&$0.9528\std{0.032}$
&$2.2567\std{0.035}$
&$0.9523\std{0.031}$
&$2.1567\std{0.034}$\\
Unemploy: Cluster-RR-GAE&$\textbf{0.9569}\std{0.0419}$ 
&$\textbf{2.0816}\std{0.0218}$
&$\textbf{0.9517}\std{0.0313}$
&$\textbf{2.0534}\std{0.0367}$
&$\textbf{0.9523}\std{0.0369}$ 
&$\textbf{2.0480}\std{0.0190}$
&$\textbf{0.9523}\std{0.0448}$  
&$\underline{\textbf{1.9503}}\std{0.0312}$\\ \midrule
Twitch: CF-GNN&$\textbf{0.9524}\std{0.0443} $ 
&$2.6634\std{0.0365}$
&$0.9523\std{0.0392}$
&$2.6835\std{0.0394}$
&$0.9529\std{0.0257} $ 
&$2.5409\std{0.0404}$
&$0.9515\std{0.0275}$
&$2.6243\std{0.0460}$\\
Twitch: Cluster-GNN&$0.9531\std{0.039}$
&$2.5894\std{0.042}$
&$0.9528\std{0.037}$
&$2.5321\std{0.040}$
&$0.9534\std{0.034}$
&$2.4892\std{0.038}$
&$0.9523\std{0.033}$
&$2.4723\std{0.041}$\\
Twitch: RR-GAE&$0.9539\std{0.036}$
&$\textbf{2.4987}\std{0.039}$
&$\textbf{0.9532}\std{0.035}$
&$2.4567\std{0.037}$
&$0.9541\std{0.032}$
&$2.3721\std{0.036}$
&$0.9529\std{0.031}$
&$2.3256\std{0.038}$\\
Twitch: Cluster-RR-GAE&$0.9515\std{0.0367}$ 
&${5.0491}\std{0.0513}$
&$0.9541\std{0.0284}$
&$\underline{\textbf{2.1005}}\std{0.0189}$
&$\textbf{0.9571}\std{0.0219}$ 
&$\textbf{2.2398}\std{0.0225}$
&$\textbf{0.9535}\std{0.0280}$
&$\textbf{2.1353}\std{0.0262}$\\ \bottomrule

\end{tabular}    
    \end{adjustbox}    
\end{table*}
\section{Other Ablation Studies}

\begin{table*}[ht]
\caption{Node Classification Results with Conformal Baselines (Coverage $\uparrow$ / Inefficiency $\downarrow$)}
\label{tab:core_results-2}
\centering
\begin{adjustbox}{width=0.95\textwidth}
\begin{tabular}{|l|c|c|c|c|c|c|c|c|}
\toprule
Dataset & \multicolumn{2}{c|}{CF-GNN [1]} & \multicolumn{2}{c|}{DAPS [2]} & \multicolumn{2}{c|}{RR-GNN (Ours)} & \multicolumn{2}{c|}{Cluster-RR-GNN (Ours)} \\ 
\cmidrule(lr){2-3} \cmidrule(lr){4-5} \cmidrule(lr){6-7} \cmidrule(lr){8-9}
Model & Cover & Ineff & Cover & Ineff & Cover & Ineff & Cover & Ineff \\
\midrule
\textbf{Cora} & & & & & & & & \\
GraphSAGE & $0.9456\std{0.0569}$ & $1.6284\std{0.0483}$ & $0.9453\std{0.0535}$ & $1.8025\std{0.0421}$ & $0.9460\std{0.0542}$ & $1.6100\std{0.0415}$ & $\textbf{0.9463}\std{0.0509}$ & $\textbf{1.6076}\std{0.0397}$ \\
SGC & $0.9461\std{0.0603}$ & $1.6633\std{0.04415}$ & $0.9452\std{0.0538}$ & $1.7856\std{0.0426}$ & $0.9462\std{0.0581}$ & $1.6297\std{0.0428}$ & $\textbf{0.9490}\std{0.0662}$ & $\textbf{1.5907}\std{0.0465}$ \\
GCN & $0.9473\std{0.0556}$ & $1.6344\std{0.0418}$ & $0.9435\std{0.053}$ & $1.7120\std{0.0354}$ & $0.9432\std{0.0573}$ & $1.6251\std{0.0367}$ & $\textbf{0.9476}\std{0.0732}$ & $\textbf{1.6315}\std{0.0303}$ \\
GAT & $0.9464\std{0.0702}$ & $1.6278\std{0.0334}$ & $0.9480\std{0.065}$ & $1.7052\std{0.0384}$ & $0.9475\std{0.0624}$ & $1.6146\std{0.0351}$ & $\textbf{0.9508}\std{0.0539}$ & $\textbf{1.6114}\std{0.0396}$ \\
\midrule
\textbf{DBLP} & & & & & & & & \\
GraphSAGE & $0.9501\std{0.0523}$ & $1.5723\std{0.0683}$ & $0.9500\std{0.0420}$ & $1.6436\std{0.0627}$ & $0.9499\std{0.0531}$ & $1.5351\std{0.0473}$ & $\textbf{0.9503}\std{0.0510}$ & $\textbf{1.5607}\std{0.0487}$ \\
SGC & $\textbf{0.9451}\std{0.0617}$ & $1.4274\std{0.0416}$ & $0.9427\std{0.0526}$ & $1.6020\std{0.0317}$ & $0.9462\std{0.0528}$ & $1.4286\std{0.0541}$ & $0.9443\std{0.0462}$ & $\textbf{1.3921}\std{0.0624}$ \\
GCN & $\textbf{0.9473}\std{0.0596}$ & $1.5644\std{0.0733}$ & $0.9458\std{0.0565}$ & $1.6384\std{0.0703}$ & $0.9458\std{0.0702}$ & $1.5512\std{0.0295}$ & $0.9430\std{0.0713}$ & $\textbf{1.5491}\std{0.0278}$ \\
GAT & $0.9467\std{0.0717}$ & $1.5729\std{0.0463}$ & $0.9455\std{0.0685}$ & $1.6493\std{0.0455}$ & $0.9485\std{0.0589}$ & $1.5725\std{0.0349}$ & $\textbf{0.9505}\std{0.0539}$ & $\textbf{1.5570}\std{0.0322}$ \\
\midrule
\textbf{CiteSeer} & & & & & & & & \\
GraphSAGE & $0.9528\std{0.0203}$ & $1.1680\std{0.0439}$ & $0.9501\std{0.0195}$ & $1.3425\std{0.0412}$ & $0.9538\std{0.0853}$ & $1.1621\std{0.0552}$ & $\textbf{0.9540}\std{0.0926}$ & $\textbf{1.1679}\std{0.0605}$ \\
SGC & $0.9525\std{0.0257}$ & $1.1827\std{0.0552}$ & $0.9513\std{0.0245}$ & $1.3578\std{0.0525}$ & $0.9579\std{0.0536}$ & $1.1782\std{0.0415}$ & $\textbf{0.9594}\std{0.0582}$ & $\textbf{1.1898}\std{0.0399}$ \\
GCN & $0.9496\std{0.0392}$ & $1.2310\std{0.0332}$ & $\textbf{0.9520}\std{0.036}$ & $1.4026\std{0.0327}$ & $0.9512\std{0.0358}$ & $1.2189\std{0.0276}$ & $0.9518\std{0.0373}$ & $\textbf{1.2153}\std{0.0290}$ \\
GAT & $0.9508\std{0.0309}$ & $1.2396\std{0.0416}$ & $0.9513\std{0.0291}$ & $1.4152\std{0.039}3$ & $0.9535\std{0.0447}$ & $1.2085\std{0.0361}$ & $\textbf{0.9562}\std{0.0491}$ & $\textbf{1.1418}\std{0.0392}$ \\
\bottomrule
\end{tabular}
\end{adjustbox}
\end{table*}

\subsubsection*{2. Stability Validation}
We systematically evaluate clustering stability and noise robustness. Results can be seen in table~\ref{tab:stability}. In this table titles, Ori means original setting. NE means noise edge. FN means feature noise. 

\begin{table}[H]
\caption{Clustering Stability Metrics (Cora Dataset)}
\label{tab:stability}
\centering
\begin{tabular}{lccc}
\toprule
Metric & Ori & +20\% NE & +15\% FN \\
\midrule
Adjusted Rand Index & 0.85 & 0.82 & 0.79 \\
Coverage Variance & 0.012 & 0.015 & 0.018 \\
Cluster Purity & 0.92 & 0.89 & 0.85 \\
Calibration Time (s) & 42.3 & 45.1 & 47.8 \\
\bottomrule
\end{tabular}
\end{table}

%\subsection*{3. Perturbation Analysis}
%\begin{table}[ht]
%\caption{Performance Under Graph Corruptions}
%\label{tab:corruption}
%\centering
%\begin{tabular}{lccccc}
%\toprule
%Corruption Type & Level & Coverage $\Delta$ & Ineff $\Delta$ & ARI & Runtime $\Delta$ \\
%\midrule
%Random Edge Add & 30\% & -1.2\% & +3.4\% & 0.75 & +7\% \\
%Feature Noise & $\sigma=0.3$ & -2.1\% & +5.7\% & 0.68 & +12\% \\
%Edge Removal & 40\% & -3.8\% & +9.2\% & 0.61 & +18\% \\
%Adversarial Attach & 15\% & -4.5\% & +11.3\% & 0.57 & +22\% \\
%\bottomrule
%\end{tabular}
%\end{table}

\end{document}